\documentclass[sn-mathphys-num]{sn-jnl}

\usepackage{graphicx}
\usepackage{times}
\usepackage{float}
\usepackage{booktabs} 
\usepackage{amsmath,amssymb,amsfonts}%
\usepackage{amsthm}%
\usepackage{mathrsfs}%
\usepackage{caption}
\usepackage{url} 
\usepackage{multirow}%
\usepackage{xcolor}%
\usepackage{textcomp}%
\usepackage{manyfoot}%
\usepackage{booktabs}%
\usepackage{algorithm}%
\usepackage{algorithmicx}%
\usepackage{algpseudocode}%
\usepackage{listings}%
\usepackage{shortbold}
\usepackage{mathptmx}

\newtheorem{theorem}{Theorem}

\newtheorem{lemma}{Lemma}

\makeatletter
\newenvironment{figurehere}
{\def\@captype{figure}}
{}
\makeatletter

\usepackage[normalem]{ulem}
\usepackage{multirow}
\usepackage{bbding}
\usepackage[numbers, sort&compress]{natbib}
\usepackage[resetlabels]{multibib}

\newcommand{\ye}[1]{\textcolor{black}{#1}}

\begin{document}

\title{\ye{Synergistic Development of Perovskite Memristors and Algorithms for Robust Analog Computing}}

\author*[1]{\fnm{Nanyang} \sur{Ye}}
\equalcont{These authors contributed equally to this work.}

\author[2,6]{\fnm{Qiao} \sur{Sun}}
\equalcont{These authors contributed equally to this work.}

\author[2]{\fnm{Yifei} \sur{Wang}}
\author[1,2]{\fnm{Liujia} \sur{Yang}}
\author[1]{\fnm{Jundong} \sur{Zhou}}
\author[5]{\fnm{Lei} \sur{Wang}}
\author[1]{\fnm{Guang-Zhong} \sur{Yang}}
\author[1]{\fnm{Xinbing} \sur{Wang}}
\author[1]{\fnm{Chenghu} \sur{Zhou}}
\author[1]{\fnm{Wei} \sur{Ren}}
\author[1,2]{\fnm{Leilei} \sur{Gu}}
\author[3,4]{\fnm{Huaqiang} \sur{Wu}}
\author*[2]{\fnm{Qinying} \sur{Gu}}\email{ynylincolncam@gmail.com; guqinying@pjlab.org.cn}

\affil[1]{\orgname{Shanghai Jiao Tong University}, \orgaddress{ \city{Shanghai}, \postcode{200240}, \country{China}}}

\affil[2]{\orgname{Shanghai Artificial Intelligence Laboratory}, \orgaddress{ \city{Shanghai}, \postcode{200232}, \country{China}}}

\affil[3]{\orgname{Institute of Microelectronics, Beijing Innovation Center for Future Chips (ICFC), Tsinghua University}, \orgaddress{\city{Beijing}, \postcode{100084}, \country{China}}}

\affil[4]{\orgname{Beijing National Research Center for Information Science and Technology (BNRist), Tsinghua University}, \orgaddress{\city{Beijing}, \postcode{100084}, \country{China}}}

\affil[5]{\orgname{Defense Innovation Institute, Academy of Military Sciences}, \orgaddress{\city{Beijing}, \postcode{100850}, \country{China}}}

\affil[6]{\orgname{Fudan University}, \orgaddress{\city{Shanghai}, \postcode{200433}, \country{China}}}

\abstract{Analog computing using non-volatile memristors has emerged as a promising solution for energy-efficient deep learning. New materials, like perovskites-based memristors are recently attractive due to their cost-effectiveness, energy efficiency and flexibility. Yet, challenges in material diversity and immature fabrications require extensive experimentation for device development. Moreover, significant non-idealities in these memristors often impede them for computing. Here, we propose a synergistic methodology to concurrently optimize perovskite memristor fabrication and develop robust analog DNNs that effectively address the inherent non-idealities of these memristors. Employing Bayesian optimization (BO) with a focus on usability, we efficiently identify optimal materials and fabrication conditions for perovskite memristors. Meanwhile, we developed ``BayesMulti", a DNN training strategy utilizing BO-guided noise injection to improve the resistance of analog DNNs to memristor imperfections. Our approach theoretically ensures that within a certain range of parameter perturbations due to memristor non-idealities, the prediction outcomes remain consistent. Our integrated approach \ye{enables use of analog computing in much deeper and wider networks, which }significantly outperforms existing methods in diverse tasks like image classification, autonomous driving, species identification, and large vision-language models, achieving up to 100-fold improvements. We further validate our methodology on a 10$\times$10 optimized perovskite memristor crossbar, demonstrating high accuracy in a classification task and low energy consumption. This study offers a versatile solution for efficient optimization of various analog computing systems, encompassing both devices and algorithms.}


\maketitle

\section*{Introduction}

The emerging field of artificial intelligence (AI), with applications like autonomous driving\citep{fayjie2018driverless}, policy optimization\citep{kang2018policy}, and complex large vision language models (LVLMs)\citep{zhou2022learning, zhang2021vinvl}, demands high-bandwidth data transfer and substantial computational resources, challenging the capacity of traditional computing systems. The separation of memory and processor in current digital computers, known as the von Neumann architecture, creates a ``memory wall" bottleneck\citep{horowitz20141, wong2015memory}. In-memory computing, leveraging nonvolatile resistive random-access memory (memristor or ReRAM), presents an effective solution by enabling direct computation within memory arrays, thus eliminating the energy-intensive and time-consuming data movement of traditional setups\citep{yang2013memristive, strukov2008missing, wang2017memristors, tuma2016stochastic, van2017non, chen2014emerging}. Employing this method, memristive crossbar arrays compute vector-matrix products—a cornerstone operation in deep neural networks (DNNs)—by encoding matrix values into memristor conductances and vector values into applied voltages\citep{yang2013memristive, ielmini2021brain}. The computational outputs are efficiently derived from the currents, in accordance with Ohm’s law and Kirchhoff’s current law \citep{yang2013memristive}, leading to analog in-memory computing characterized by significantly enhanced speed and energy efficiency\citep{joksas2022nonideality}.

Memristors, stemming from the fusion of ``memory" and ``resistor", are two-terminal passive devices capable of precise resistance modulation through electrical stimulation. They have been considered as critical components for neuromorphic computation due to their advantages of high-density integration in a cross-point array, multi-level memory, and good scalability\citep{sun2019understanding, wang2020resistive}. 
Memristors typically consist of a three-layered structure, with a switching layer sandwiched between two metallic electrodes. While this layer has primarily been constructed from inorganic materials like metal oxides\citep{yao2017direct, hu2016dot, park2013situ, yang2012observation}, amorphous silicon\citep{huang2011resistive, yao2011silicon}, and chalcogenides\citep{kozicki2006mass, wang2006resistive}, there has been a recent surge of interest in employing 
organic-inorganic hybrid perovskites (OHPs) for neuromorphic devices\citep{wang2020resistive, yang2013memristive}. 
OHPs are notable for their mixed ionic-electronic conduction, enabling low-energy halide counterion movement under electric fields, making them ideal for energy-efficient neuromorphic computing that mimics synaptic functions\citep{xiao2020recent, yoo2015resistive}. Additionally, in contrast to inorganic memristors, OHPs offer cost benefits, superior optical and charge-transport characteristics, and mechanical flexibility, positioning perovskite memristors as a promising option for future neuromorphic computing advancements.

Although memristors share remarkable similarities with biological components like synapses and neurons in both their physical mechanisms and functional behavior, it is imperative to understand that these resemblances do not inherently ensure efficient computation\citep{yang2013memristive,wang2020resistive}. This is primarily because contemporary state-of-the-art DNNs rely on digitalized values for connection weights. In contrast, representing these weights using analog memristive conductance encounters challenges characterized by non-ideal behaviors, including nonlinear conductance responses, stochastic conductance changes, device-to-device variability, programming errors, etc\citep{wan2022compute,kiani2021fully}. DNNs built on memristors (denoted as analog DNNs) are particularly susceptible to these non-idealities due to the absence of a potential gap between high and low voltages for noise resistance\citep{xiao2020analog,ye2023improving}. Consequently, the parameters of analog DNNs, represented by the conductances, can be easily distorted, potentially compromising the effectiveness of analog deep-learning systems. Many dedicated efforts have been made to address this issue, with device-algorithm co-design emerging as a promising strategy for achieving efficient analog computing\citep{ambrogio2018equivalent,sun2018ti,liurescuing,dalgaty2021situ,chen2017accelerator,stathopoulos2017multibit}. 

While extensive co-optimization has been explored in mature inorganic memristors, research on perovskite counterparts remains limited\citep{shi2021review}. Unlike inorganic electronics, which benefit from established manufacturing processes and seamless integration with existing semiconductor technologies, perovskite memristors entail diverse material options and various fabrication methods, presenting challenges in developing optimal devices and integrating suitable DNNs\citep{xiao2020recent,yoo2015resistive,misra2017low}. Furthermore, these memristors often suffer from substantial non-idealities arising from immature fabrication processes, limited material durability, and environmental sensitivity, hindering efficient DNN implementations and impeding their application in analog computing. Most efforts have concentrated on hardware modifications, such as material engineering and fabrication optimization, to address non-idealities\citep{park2021organic,kumar2018compliance,john2018ionotronic,xiao2016energy,yang2020leaky,sun2019understanding}. However, this approach is labor-intensive and time-consuming. It is worth noting that the mismatch between devices and algorithms can also be addressed from an algorithmic perspective. This involves either mitigating device non-idealities or leveraging unexpected properties as valuable features for novel computing paradigms, an area that has received relatively less attention.

\begin{figurehere}
 \begin{center}
 \includegraphics[width=0.95\linewidth]{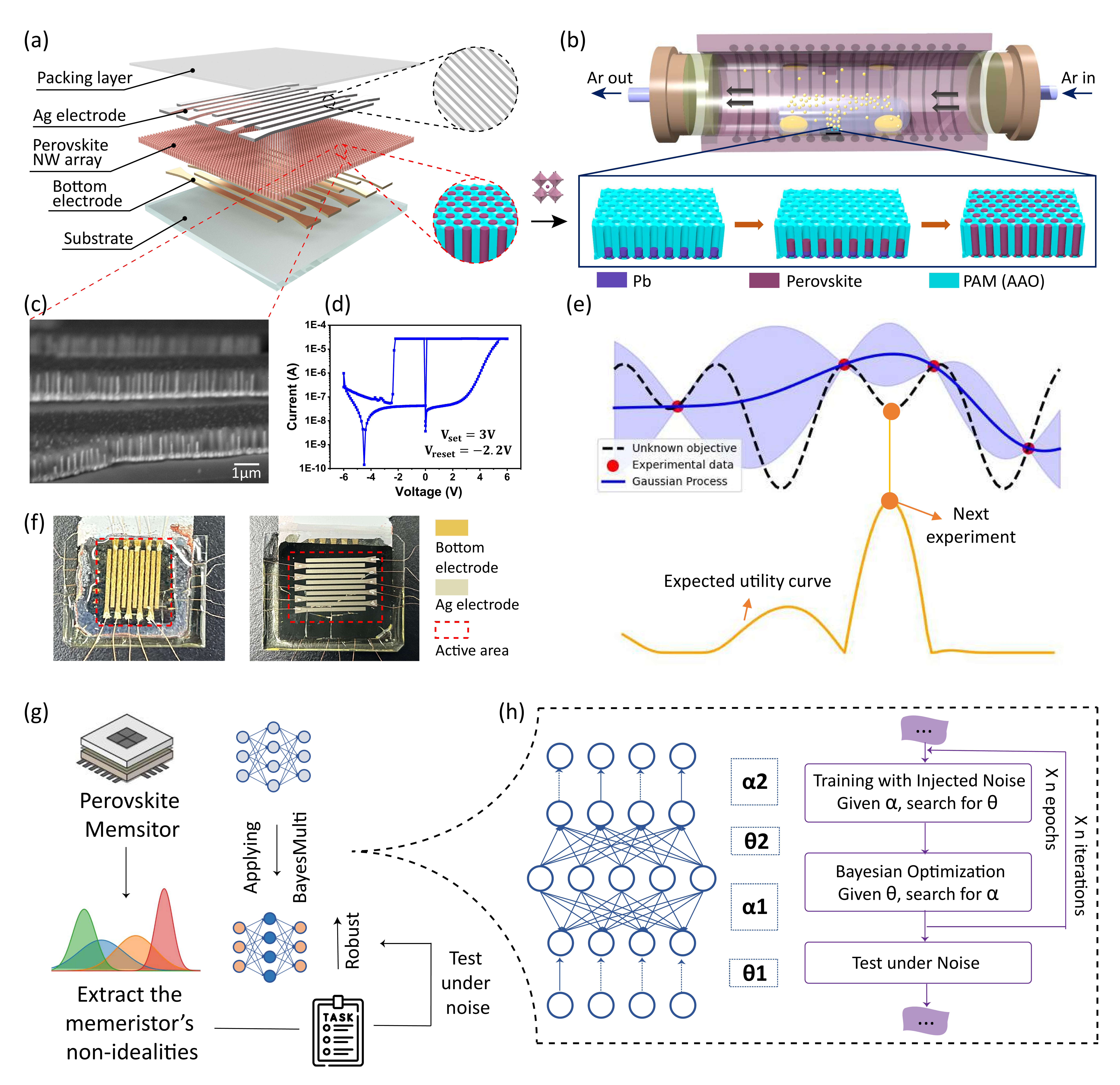}
    \caption{(a) Illustration of a perovskite NW-based memristor crossbar featuring Ag as the top electrode, a perovskite nanowire array as the switching layer, and Al or Au as the bottom electrode. (b) Depiction of perovskite nanowire synthesis via chemical vapor deposition (CVD) from an AAO PAM template; (c) Cross-sectional SEM image of $\text{MAPbBr}_3$ NWs in PAM. (d) I-V characteristic of Ag/$\text{MAPbBr}_3$ NWs/Al memristor devices. (e) Schematic representation of Bayesian optimization in perovskite memristor fabrication, utilizing a Gaussian process surrogate model informed by experimental data to guide experimental design, illustrated by a mean function with uncertainty indicated by the shaded area. (f) Photos of a 10$\times$10 perovskite NW-based memristor crossbar. (g) The workflow of Bayesian noise injection optimization. Memristors' non-idealities are extracted from a series of memristors, acting as hardware noise for testing. BayesMulti is applied to different models/networks for specific tasks to improve the robustness. Different Hardware noise is applied as perturbations to these models/networks to validate the effectiveness of Bayesmulti and the guidance of hardware optimization. (h) Detailed framework for implementing BayesMulti.}
    \label{fig: overview}
 \end{center}
\end{figurehere}

In this study, we pioneer a novel approach by co-developing suitable perovskite nanowire (NW)-based memristors and robust analog DNNs capable of counteracting the inherent non-idealities of these memristors. This integration yields analog DNNs that are adaptable to complex computational tasks. We introduce the concept of ``usability", derived from basic current-voltage (I-V) characteristics of perovskite memristors, to evaluate their suitability for analog computing under specific fabrication conditions. Leveraging Bayesian optimization (BO) with ``usability" as the target, we efficiently identify optimal material selection and fabrication conditions within a vast search space, significantly reducing time consumption compared to human-driven approaches. Concurrently, we develop a noise injection approach, named ``BayesMulti", also guided by BO to optimize the characteristic of the injected noises with multinomial distributions, thereby enhancing the robustness of analog DNNs built upon these memristors. We further provide theoretical proof that BayesMulti ensures consistent prediction outcomes within a specified parameter perturbation range due to memristor non-idealities. Our algorithmic framework outperforms state-of-the-art methods across various tasks, from object detection to LVLMs, and various hardware non-idealities. With optimized device and algorithms in hand, we further conducted a classification task in a 10$\times$10 perovskite memristor crossbar, achieving high accuracy and reduced performance decline compared to digital DNNs. This underscores the fault-tolerant, noise-robust, and generalizable nature of our analog neural networks to complex models and tasks.

\section*{Results}
\subsection*{Bayesian Fabrication Optimization for Perovskite Memristors}
The perovskite nanowire (NW)-based memristor structure is depicted in Figure \ref{fig: overview}a. Lead halide perovskites ($\text{APbX}_3$; A=methylammonium, formamidinium, cesium; X=I, Br, Cl) are grown inside nanoengineered porous alumina membrane (PAM) pores through a bottom-up chemical vapor-solid-solid reaction (VSSR), resulting in three-dimensionally integrated perovskite NWs (Figure \ref{fig: overview}b). As shown in Figure \ref{fig: overview}c of the cross-sectional SEM image, these NWs are embedded within the PAM, serving as the switching layer and positioned between the active electrode (Ag) and counter electrode (Al or Au). Figure \ref{fig: overview}d displays a typical I-V characteristic of a perovskite NW-based memristor. A DC voltage sweep from 0 → + 6 V → 0 V → $-6$ V → 0 V was applied with compliance current of $2\times10^{-5}$ A to prevent breakdown.` The device demonstrated distinct SET and RESET events at threshold switching voltages of 3 V and -2.2 V respectively. The resistive switching behavior arises from the electrochemical metallization (ECM) of Ag, leading to the formation of inhomogeneous filaments. The fundamental structure and switching mechanism have been previously reported\citep{waleed2017lead, gu20163d, zhang2021three}. These works have also presented that perovskite NW-based memristors can achieve multilevel resistance states and excellent data fidelity. Additionally, the unique NW array structure, coupled with lateral passivation of the PAM, has improved both material and electrical stability in the devices, leading to enhanced cycle endurance and retention capability\citep{zhang2021three}. Figure \ref{fig: overview}f presents a 10$\times$10 memristor crossbar fabricated from the perovskite NW array. The scalability and ease of fabrication of these devices underscore their potential for significant applications in data storage and neuromorphic computing.

The switching performance of perovskite NW-based memristors can be influenced by various factors, including perovskite types, PAM morphology, electrodes, etc. While previous research focused on understanding memristor mechanisms, achieving minimal non-idealities is crucial for neuromorphic computing applications in this work. However, this optimization is multidimensional, making manual evaluation impractical due to resource constraints. Therefore, we employ Bayesian optimization (BO), an effective method for optimizing expensive functions, which has demonstrated superiority over other approaches\citep{liang2021benchmarking,griffiths2020constrained,shields2021bayesian}, to tackle this issue. BO efficiently balances uncertainty exploration and information exploitation, to high-quality configurations with fewer evaluations. Notably, BO is versatile and applicable to diverse search spaces\citep{shahriari2015taking,snoek2012practical,frazier2018tutorial}.

To establish an efficient and precise BO framework, we have to first define the search space and optimization objectives. Based on prior research and our expertise, we have identified five influential factors as variables for memristor optimization: (1) perovskite types, (2) NW length, (3) NW diameter, (4) lead electrodeposition (Pb ED) time, and (5) Ag thickness. These collectively constitute the final search space, encompassing 8400 possible experimental configurations. Additional information on search space selection is available in Supplementary Note 2. The optimization target is to assess a typical perovskite memristor's suitability for analog computing. Hence, our primary aim is to determine how to quantify this ``suitability".

Considering an analog DNN as a series of nonlinear functions, the goal is to minimize the loss $\ell\left(f_{\theta}(x), y\right)$ for input data $x \in \mathcal{R}^{d}$ and its corresponding label $y$, where $\ell$ represents the loss function and $f_{\theta}$ is the neural network parameterized by weights $\theta$. Device imperfections, such as conductance instability, variations across cycles/devices, and programming errors, can cause weight parameters $\theta$ to drift, affecting the DNN's robustness and accuracy. Therefore, assessing these non-idealities based on memristor characteristics is vital in determining a memristor's aptness for analog computing. We categorize these non-idealities into two types: non-monotonic non-ideality and stochastic non-ideality.\\
\textbf{Non-monotonic Non-ideality} Ideally, the conductance of a memristor would exhibit a proportional relationship with input electric charge, allowing for straightforward conductance control through charge modulation without the requirement for complex circuitry. However, in practice, even with a forward charge, the conductance may not exhibit a strictly ascending trend. This therefore induces non-monotonic non-ideality to the neural network parameters. \\
\textbf{Stochastic Non-ideality} In practice, variations in conductance between measurements are inevitable due to environmental disturbances and inherent non-idealities, such as cycle and device variations, and thermal and electrical noises within the device. This introduces stochastic non-ideality to the neural network. 

To emulate the two types of non-idealities, we conducted simulations based on a memristor's electrical characteristics, as detailed in the Methods section. This approach offers a straightforward and fundamental means of assessing the performance of a typical device, making it widely applicable across different material systems and accessible to researchers lacking advanced experimental facilities. Based on the simulations, we introduce the concept of ``usability" to assess a typical memristor's suitability for analog computing. ``Usability" is employed as the optimization objective for the subsequent BO process, serving as an indicator for the prosperity of a specific perovskite manufacturing design. Additionally, by extracting device non-idealities, we developed \textbf{PerovskiteMemSim}, detailed in Supplementary Note 4, for further evaluation of analog computing systems, including both devices and algorithms, across various applications.

\begin{figure}[H]
    \centering
    \includegraphics[width=\linewidth]{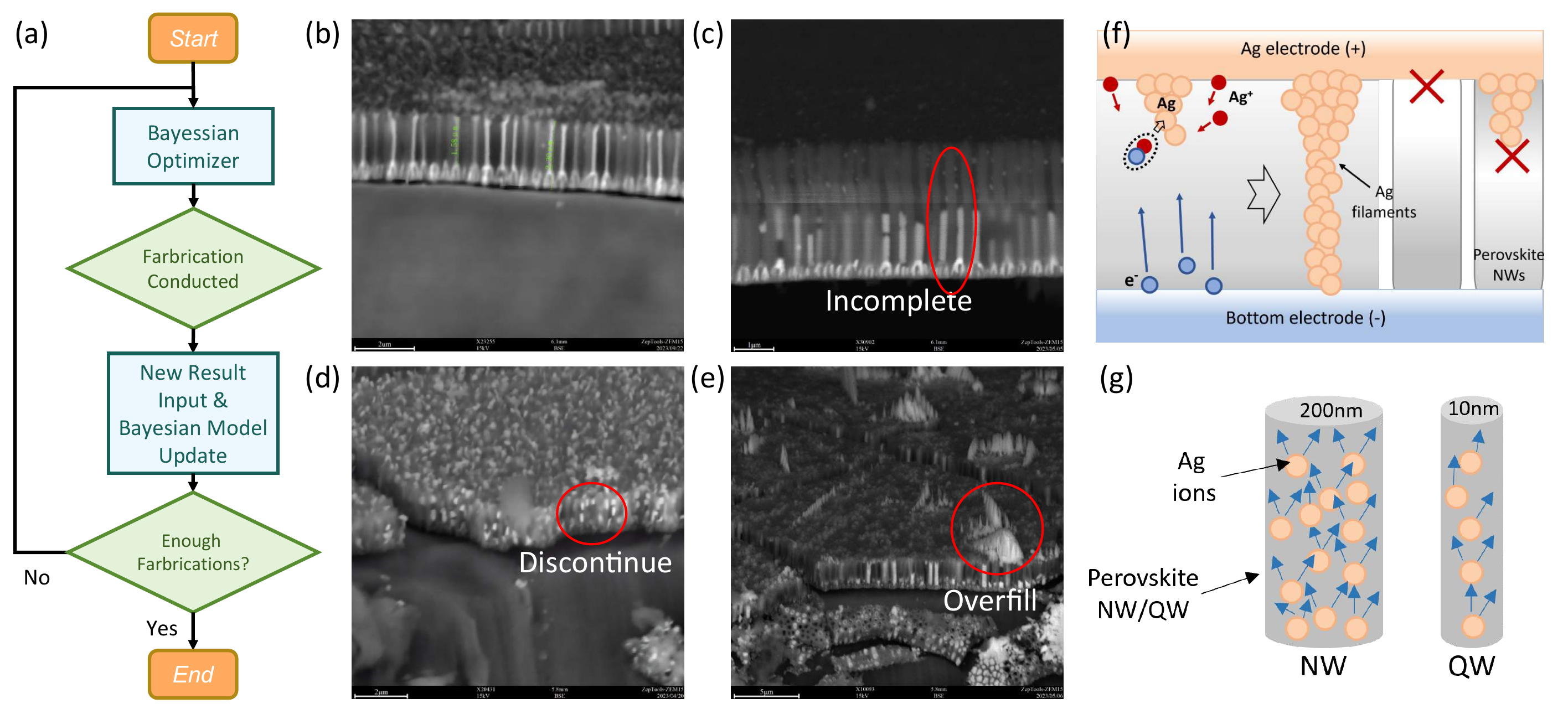}
    \caption{(a) Illustration of the Bayesian fabrication optimization process. (b) SEM images from a good-performed memristor, poorly-performed memristor with incomplete(c), discontinued (d), and excessive growth of perovskite (e). (f) A uniform and completely grown of perovskite inside nanowires ensures the formation of Ag filament while in other irregular cases, filament formation and rupture are interfered. (g) Schematic showing differences in the number of electron movement pathways in NWs and QWs.}
    \label{fig: SEM}
\end{figure}

\subsubsection*{Bayesian Fabrication Optimization Process} 
For a given search space, Bayesian optimization begins by gathering initial outcome data from experiments or existing literature. These data are then used to train a probabilistic surrogate model, which is constructed by combining previous observations with prior functions that allow for inference of globally optimal locations. After training the surrogate model, new experiments in the search space are selected sequentially by optimizing an acquisition function that maximizes the observed metric scores of candidate experiments for the next evaluation. Subsequently, the proposed experiments are conducted, their results are added to the record, and the posterior of the surrogate model is updated. This process of surrogate modeling and acquisition function maximization iterates, as shown in Figure \ref{fig: overview}e, effectively guiding navigation through the search space and ultimately leading to convergence towards optimal or near-optimal configurations\citep{shahriari2015taking}. 

Having established the BO architecture, our focus shifts to optimizing its performance through critical algorithmic component designs. These include encoding methods, surrogate model types, and acquisition function selections, which play a pivotal role in maximizing usability in optimization\citep{shahriari2015taking}. The aim of BO in this study is to optimize the usability of memristors across a combinatorial set of thousands of possible fabrication and characterization conditions. We employ numerical encoders to convert text variables (e.g., material selections) into structured representations for processing by BO.
The surrogate model's primary requirement is to make predictions and estimate variance. Inspired by prior successful applications in Bayesian reaction optimization\citep{shields2021bayesian}, we opt for the Gaussian Process (GP) as the surrogate model. The GP model is characterized by the mean function and the covariance function, which guide the acquisition function derivation. Once the surrogate model is trained, new experiments in the search space are selected sequentially through acquisition function optimization. To strike a balance between exploration and exploitation, we employ Expected Improvement (EI) as the acquisition function. Maximizing EI leads to the selection of candidate configurations expected to outperform the current best results significantly. The selected experiment is then conducted, and its corresponding usability is calculated and integrated into the BO process to update the surrogate model's posterior. This process continues iteratively until usability is maximized, resources are depleted, or further improvements are improbable. Further details on the BO process for memristor fabrication optimization are outlined in Supplementary Note 2.

In our study, we commenced the BO process with an initial experiment guided by prior research findings and the capabilities of our laboratory. This initial setup, featuring methylammonium lead tribromide ($\text{MAPbBr}_3$) as the perovskite-type, 1 $\mu m$ NW length, 250 nm NW diameter, a 10-minute lead electrodeposition (Pb ED) period, and an Ag thickness of 100 nm, achieved a usability score of 0.36. We numerically encoded the perovskite type, a discrete variable, and treated others as continuous variables. Starting from this point, BO iteratively updates the surrogate model posterior through human-in-the-loop experimental outcomes integration, as depicted in Figure \ref{fig: SEM}a (fabrication details are in the Method section). Notably, BO efficiently explored the entire search space, avoiding local maxima and converging to optimal experimental conditions in just twelve iterations. The chosen experiments by the BO process and the corresponding usability values are outlined in Figure~\ref{fig:BO fabricate}. The optimal condition decoded as $\text{MAPbI}_3$ perovskite-type, 1.5 $\mu m$ NW length, 150 nm pore diameter, 15-minute Pb ED, and a Ag thickness of 200 nm, resulted in a significantly improved usability score of 0.93. It's worth highlighting that this top-performing configuration involved unconventional parameters, such as smaller NW diameters. This underscores BO's ability to discern quantitative conditions in human unexplored regions of the search space. 

BO has demonstrated its effectiveness in guiding the fabrication of perovskite NW-based memristors for maximizing usability. However, given the highly complex nature of DNNs, achieving optimal performance necessitates a coordinated optimization of both material selection and DNN architectures. This involves two primary objectives: (1) developing a training methodology for analog DNNs that are resilient to device non-idealities, a topic we will delve into in the following section, and (2) creating a range of memristors that uniformly span the entire usability spectrum (i.e. 0 to 1). This enables precise, quantitative evaluation of the enhanced robustness of analog DNNs against various non-idealities. By adjusting the optimization objective in the BO process, we obtained a series of experimental configurations allowing us to fabricate perovskite NW-based memristors with different usabilities ranging from 0.12 to 0.93, as illustrated in Figure \ref{fig:usability_distribution} showing the usability distribution. 

To investigate performance variability among memristors, we conducted a comparative micro-structural analysis. Scanning electron microscope (SEM) images in Figure \ref{fig: SEM}b illustrate that well-performing memristors feature nanowires uniformly and completely grown with perovskite. This uniform growth is crucial for enabling the movement of Ag ions and the formation of Ag filaments, which are essential for resistive switching in these devices. Conversely, the poorly-performing memristors, as depicted in Figure \ref{fig: SEM}c-e, exhibit irregularities in perovskite growth within the nanowires, such as incomplete, discontinued, or excessive growth. These irregularities likely interfere with filament formation and rupture, impacting the overall performance of the memristors. Our findings also indicate that memristors based on formamidinium (FA) and cesium (Cs) consistently exhibit lower usability compared to those using methylammonium (MA) perovskite. This discrepancy is likely due to the FA-based memristors inevitably undergoing an unwanted $\alpha$-to-$\delta$ phase transition and the challenges in controlling the VSSR process for Cs-based memristors. Additionally, we discovered that memristors with medium pore diameters generally perform better, attributed to a higher number of residual Ag paths facilitating electron movement in larger-diameter nanowires compared to quantum wires \citep{poddar2021down}, as illustrated in Figure \ref{fig: SEM}g. However, when pore diameters exceed 250 nm, performance declines, possibly due to the complexities of controlling the nanowires' pore enlargement treatment.

\subsection*{Bayesian Noise Injection Optimization for Robust Analog DNNs}
As mentioned, achieving optimal performance of memristor-based analog DNNs requires a synergistic optimization of both device fabrication and DNN architectures. Therefore, our next aim is to develop a training strategy to enhance the robustness of analog DNNs against memristor non-idealties. Previous research has highlighted inductive noise as a crucial architectural factor in DNNs\citep{ye2023improving}. By implementing noise injection, we enable random parameter space sampling in analog DNNs, moving beyond the limitations of traditional single-point optimization. This approach effectively widens the robust parameter space, ensuring algorithmic stability within a permissible perturbation range. In essence, if the randomized analog DNN provides a correct prediction, it will maintain the same performance for any perturbation within this range. This suggests that analog DNNs can be equivalent to digital DNNs under certain conditions, making analog DNNs suitable for a wide spectrum of applications.

Selecting an appropriate noise spectrum is critical as it determines the sampling positions around the original parameter point and, consequently, the radius of the robust region of analog DNNs. From our analysis of potential spectral types (Multinomial, Gaussian and Laplace), we find that the multinomial distribution generates a substantial perturbation range within which the analog DNN remains robust. The distribution is as follows:

\begin{center}
\begin{equation}
\text{PDF}(\eta) = \left\{ 
  \begin{array}{ll}
    p_1 & \text{if } \eta=0, \\
    p_2 & \text{if } \eta=0.5, \\
    \max \{1-(p_1+p_2), 0\} & \text{if } \eta=1.
    
  \end{array} 
  \label{eq:multinomial}
\right. 
\end{equation}
\end{center}
where $\text{PDF}$ represents the probability density function, $p_1$ and $p_2$ are the two independent coefficients that control the multinomial distribution. We theoretically analyzed the advantage of the multinomial noise injection as shown in Theorem~\ref{thm:maintheorem}. Intuitively, the injected multinomial noise can ``immunize" the DNN to device non-idealities. We quantify the robustness of the DNN to non-idealities by the maximum allowable disturbance $\delta$ that can be applied to the network parameters. 
Given that multinomial noise offers immunity when $\delta_i=0$ and $\delta_i=0.5$, we concentrate on the maximum number of parameters allowed to change, excluding $\delta_i=0.5$ cases already ``immuned" by this noise. We used a functional optimization framework to derive Theorem~\ref{thm:maintheorem}.

\begin{theorem}
\label{thm:maintheorem}
\textbf{\textup{(Robustness guarantee of noise injection)}}. Given the analog DNN model $f$ trained with induced multinomial noises $\pi_{0}$ following the PDF in Equation~\ref{eq:multinomial}, $f_{\pi_{0}}(\theta_{0}):= \Ebb_{\eta \sim \pi_{0}}[f(\theta_{0}*\eta)]$, where $\theta_{0}$ is the DNN's parameters. The maximum allowable multiplicative disturbance $\delta$ can be applied on $\theta$ is within the robustness set $\mathcal{B}=\{ \delta : ||\delta - 1||_0 + ||\delta - 0.5||_0 - \Theta \leq r \}$, with $r$ satisfying
\begin{equation}
        r \leq \frac{\ln(1.5-f_{\pi_0}(\theta_{0}))-\Theta \ln(1-p_2)}{\ln p_1-\ln (1-p_2)}
\end{equation}
where $\Theta$ is the dimension of $\theta_{0}$, \textit{i.e.,} the number of parameters .

\end{theorem}

Theorem~\ref{thm:maintheorem} demonstrates that the parameters $p_1$ and $p_2$ of the induced multinomial noises synergistically enhance the robustness of analog DNNs, indicated by a larger robustness set radius $r$. Increasing $p$ values, representing higher noise levels can theoretically improve robustness but may reduce accuracy due to training challenges under significant noise. Compared with increasing $p_1$, increasing $p_2$ brings milder improvement on the robustness and less difficulty in training. Balancing robustness and accuracy is crucial due to these complex effects. To automatically optimize the noise injection settings to ensure robustness without compromising accuracy, we leveraged Bayesian optimization. We introduced noise injection layers after each DNN layer, excluding the final softmax output layer. We denote the specification of the additional noise injection layers as $\alpha$. Given the absence of exact gradient information for $\alpha$, we employed Bayesian optimization with a Gaussian Process surrogate model to search for the optimal $\alpha$ within the search space (Figure \ref{fig: overview} f-g). We named our method ``BayesMulti" and the detailed theoretical proof and BO process are presented in Supplementary Note 3.

\subsection*{Discussion on the Effectiveness of Usability and BayesMulti}
Having optimized perovskite NW-based memristors and developed a fault-tolerant training method for robust analog DNNs, our next objective is to validate the suitability of these memristors for analog computing, assessing whether usability can serve as an indicator for fabricating desirable memristors. Concurrently, we aim to confirm the effectiveness of BayesMulti across diverse tasks under varying levels of memristor non-idealities.

For validation, we chose tasks that span a broad spectrum of model complexities and applications. These include image recognition, autonomous driving, antigen-antibody matching, and large vision and language models (LVLMs). The datasets employed for these tasks are diverse, encompassing the Modified National Institute of Standards and Technology (MNIST)\citep{deng2012mnist}, Canadian Institute for Advanced Research-10 (CIFAR-10)\citep{krizhevsky2010cifar}, Karlsruhe Institute of Technology and Toyota Technological Institute at Chicago Vision Benchmark (KITTI)\citep{geiger2013vision}, Human Immunodeficiency Virus (HIV)\citep{osti_1458915}, Coronavirus Antibody Database (CoVAbDab) SARS-CoV-2\citep{raybould2021cov}, and MiniGPT4 datasets\citep{zhu2023minigpt}. To evaluate the effectiveness of BayesMulti, we implemented the state-of-the-art Empirical Risk Minimization (ERM) algorithm\citep{zhang2017mixup} as a baseline method, which focuses solely on minimizing the empirical risk, for comparison. Each method was executed ten times under different levels of non-idealities using \textbf{PerovskiteMemSim}, and the mean (dot) and standard deviation (shaded area) of evaluation metrics (e.g. accuracy) were recorded. The implementation details of the training method (i.e. BayesMulti) and the test conditions (i.e. different hardware noise derived from real perovskite memristors) are presented in Supplementary Note 4.

\subsubsection*{Evaluations on image classification}
\begin{figure}[H]
    \centering
    \includegraphics[width=\linewidth]{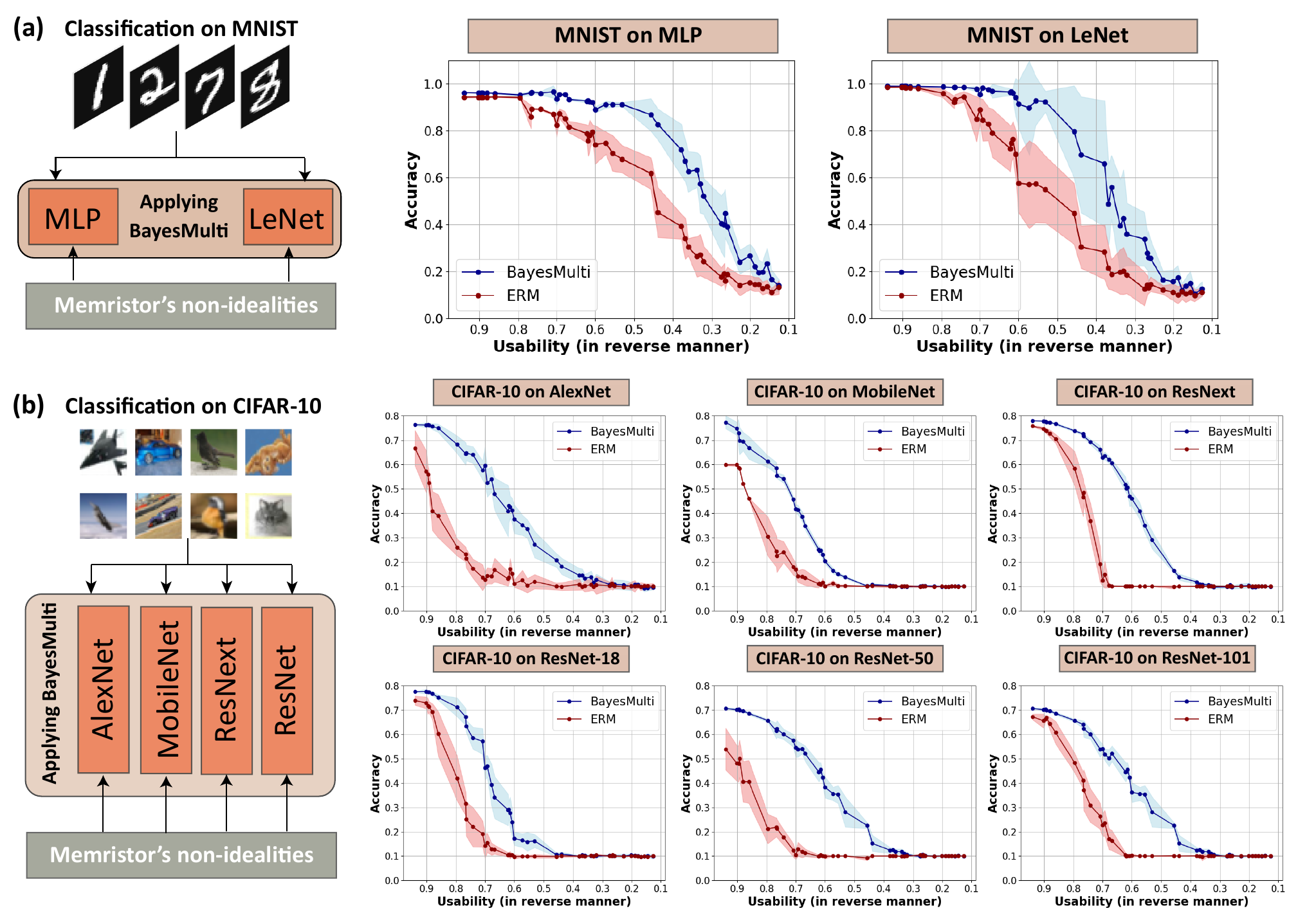}
    \caption{(a) Experimental results on the MNIST dataset. Left: a schematic demonstration of the task. Right: the curve charts compare the prediction accuracy of BayesMulti and ERM at different usability levels on MLP and LeNet. (b) Experimental results on the CIFAR-10 dataset. Left: a schematic demonstration of the task. Right: the curve charts compare the prediction accuracy of BayesMulti and ERM at usability levels on AlexNet, MobileNet, ResNext, and ResNet. Each method was run 10 times, and the mean (dot) and standard deviation (shaded areas) of accuracy under different usability levels are recorded and demonstrated in the curve charts.}
    \label{fig: image cls and ASTE}
\end{figure}

Performance validation was first conducted on the MNIST dataset (Figure \ref{fig: image cls and ASTE}a), utilizing a three-layer multilayer perceptron (MLP)\citep{murtagh1991multilayer} and a LeNet5 network\citep{lecun1998gradient}, with usability ranging from 1 to 0.1 for these models. The results indicate that BayesMulti surpasses ERM across the entire device usability spectrum, achieving significantly higher accuracy, particularly at lower usability levels of 0.5. For classification tasks in Lenet5, ERM exhibits marked accuracy deterioration at usability of approximately 0.8, while the accuracy of BayesMulti remains relatively consistent within a variance region of usability$\leq 0.6$. This trend is similarly evident in MLP, where BayesMulti demonstrates superior noise robustness compared to ERM.

We then employed a consistent experimental methodology across various neural network architectures on the CIFAR-10 dataset, renowned for its real-world object recognition challenges compared to the simpler MNIST dataset's handwritten digits. Our findings, detailed in Figure \ref{fig: image cls and ASTE}b, reveal that BayesMulti surpasses the baseline across all network architectures and noise levels. Specifically, under a usability level of 0.7, BayesMulti maintains a robust accuracy above 0.6 in AlexNet and ResNeXt, whereas the ERM accuracy plummets to mere chance (0.1). While both methods exhibit diminished performance on MobileNet, BayesMulti consistently outperforms ERM, particularly at higher usability levels (i.e. usability$\geq 0.65$), achieving more than double the ERM's performance in some cases. Additionally, BayesMulti demonstrates great performance across varying depths of ResNet, with notable advantages in ResNet-50 over ResNet-101. This suggests a nuanced interplay between network depth and performance, where increased model complexity does not necessarily equate to improved accuracy.

\subsubsection*{Evaluations on autonomous driving}
\begin{figure}[H]
    \centering
    \includegraphics[width=1\linewidth]{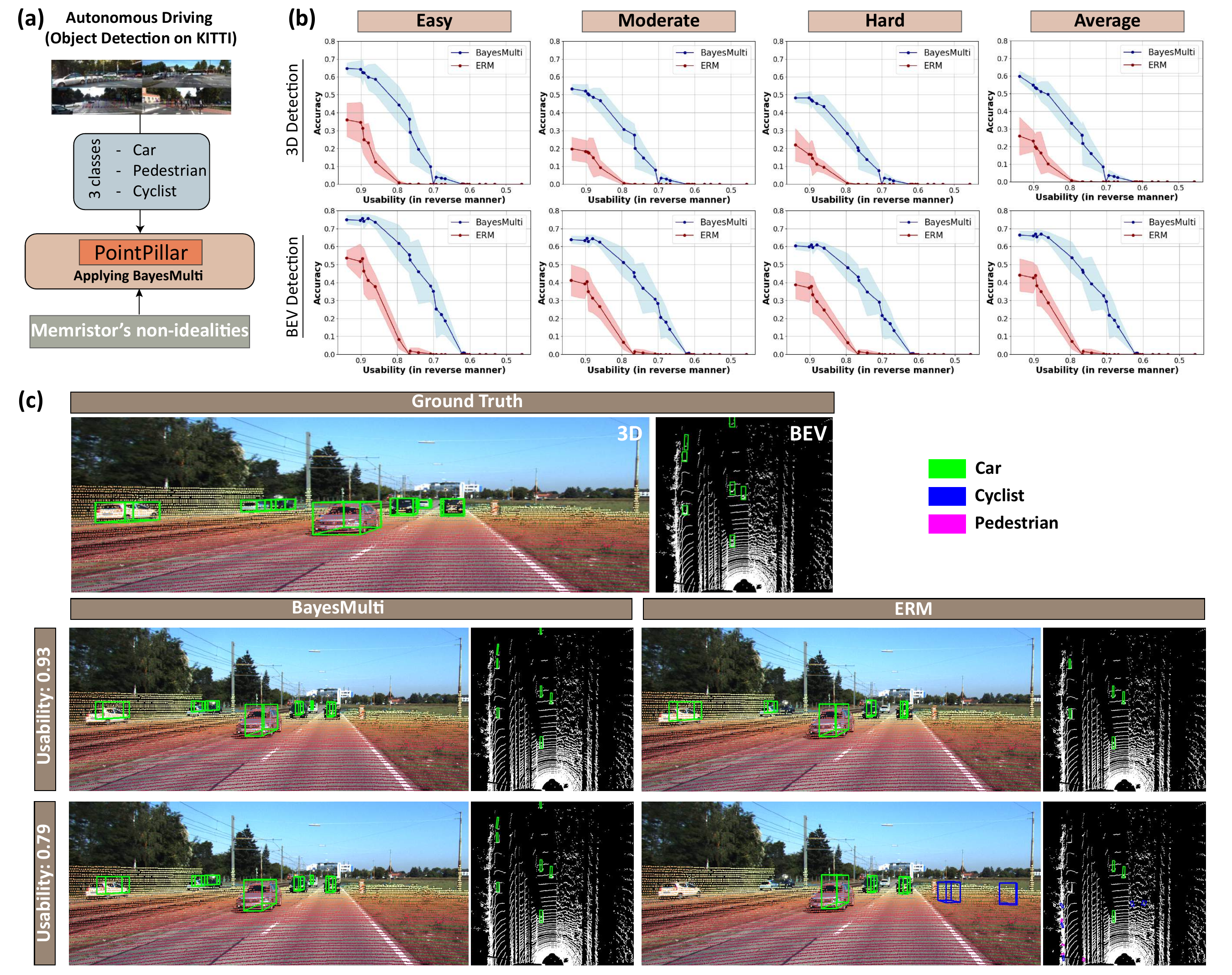}
    \caption{(a) The experimental setting of the object detection task on KITTI. Three types of objects were detected: cars, pedestrians, and cyclists. (b) Detection accuracy of BayesMulti and ERM on all KITTI dataset subsets (Easy, Moderate, and Hard) together with the average performance. Each method was run 10 times, and the mean (dot) and standard deviation (shaded areas) of accuracy under different usability levels are recorded and demonstrated in the curve charts. (c) Visualization of object detection results. The top figure is the 3D and BEV of the ground truth detection result. The left bottom figure is BayesMulti's result and the bottom right figure is ERM’s result under different analog noise levels.}
    \label{fig: autodriving}
\end{figure}

To further assess the effectiveness of BayesMulti under more stringent conditions, we extended our experiments to the task of point cloud detection for autonomous driving—a sector where precision is critical. Point clouds, representing spatial data in three dimensions, are pivotal to the functionality of modern autonomous vehicles\citep{guo2020deep}. This segment of computer vision is particularly demanding, necessitating extensive datasets and intricate models to ensure safety and efficiency\citep{gupta2021deep}, thereby underscoring the potential utility of ReRAM devices. Nonetheless, the propensity for weight-shifting biases to accumulate during extended forward propagation presents a significant challenge.
Our investigation utilized the widely-used Velodyne Lidar dataset, KITTI \citep{geiger2013vision}, focusing on the detection of cars, pedestrians, and cyclists. We employed two established object detection metrics, Bird’s Eye View (BEV) and 3D Detection, to evaluate the performance of ERM and BayesMulti. The comparative analysis was conducted through the lens of the PointPillars network—an innovative and efficient point-cloud-based object detection algorithm tailored for autonomous driving applications \citep{lang2019pointpillars} (Figure \ref{fig: autodriving}a). Details of the model architecture and noise injection are presented in Figure \ref{fig: Archi PointPillar} and Supplementary Note 5.

Figure \ref{fig: autodriving}b illustrates the consistent superiority of BayesMulti over ERM across all levels of usability for BEV and 3D detection metrics within each subset of the KITTI dataset (easy, moderate, and hard). Notably, the performance gap between BayesMulti and ERM here is more pronounced compared to other tasks previously mentioned, even at high usability levels. For example, at usability$\textgreater 0.9$, BayesMulti's accuracy in 3D detection scenarios is more than double that of ERM. This performance gap widens as usability decreases, with BayesMulti achieving up to 10 to 100 times the accuracy enhancement at some usability levels. When the usability decreases to 0.8, the accuracy of ERM sharply drops to zero, failing to correctly identify any objects. In contrast, BayesMulti maintains commendable accuracy, sustaining an accuracy of approximately 0.3 even under such adverse conditions. The visual representations of car detection in Figure \ref{fig: autodriving}c support these observations. Note that compared to other tasks, the sensitivity to non-idealities in PointPillar is much more significant, it is reasonable due to the model's complexity and the stringent accuracy requirements of autonomous driving tasks, which become more challenging as task difficulty increases. BayesMulti's capability in such a challenging scenario highlights its significance and potential for deploying life-critical applications in future analog computing. 


\subsubsection*{Evaluations on biological applications}
We further conducted evaluations of our method in the context of biological applications. Our first task focuses on predicting the neutralization effects of antibodies (Abs), particularly for those that have not been previously characterized through experimental interactions with antigens (Ags). Given the time-consuming and resource-intensive nature of wet lab experiments, there is a growing need in this field for fast and accurate computational methods to expedite the discovery of novel therapeutic antibodies.
For this task, we employed Mason's CNN\citep{mason2019deep, mason2021optimization}, a sequence-based model that has shown effectiveness in wet-lab experiments, to work with the HIV dataset\citep{osti_1458915} and the CoVAbDab SARS-CoV-2 dataset\citep{raybould2021cov}. We incorporated enhancements proposed in previous studies\citep{zhang2022predicting} into Mason's CNN architecture, which includes the addition of an Ag extraction module and an Ab-Ag embedding fusion module. These modifications allow us to construct dynamic relation graphs to quantify the relationships among Abs and Ags, addressing the original model's limitation of learning only antibody features for a single antigen. Further details regarding the network architecture and noise injection can be found in Figure \ref{fig: MasonCNN} and in Supplementary Note 5.

\begin{figure}
    \centering
    \includegraphics[width=\linewidth]{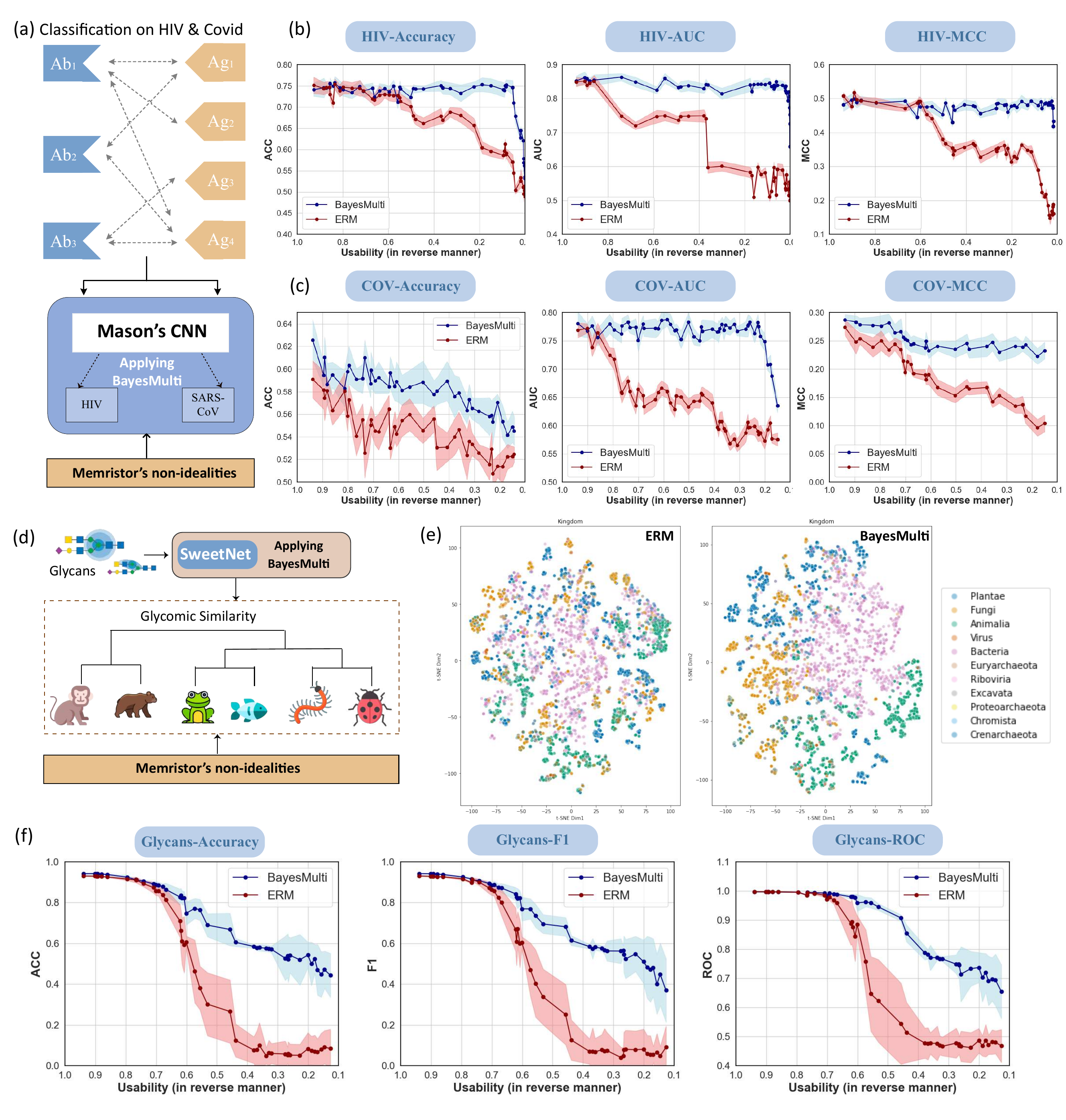}
    \caption{(a) A schematic demonstration of the task of predicting neutralization effects of Abs, with BayesMulti applied to Mason's CNN. (b) The accuracy, AUC, and MCC of BayesMulti and ERM at different hardware non-idealities (usability levels) on the HIV dataset and (c)  the CoVAbDab SARS-CoV-2 dataset. (d) A schematic demonstration of species prediction task by glycans' representations, with BayesMulti applied to the SweetNet. (e) Taxonomic glycan representations learned by SweetNet trained with ERM and BayesMulti. Glycan representations. These representations are shown via t-SNE and are colored by their taxonomic kingdom. (f) The accuracy, F1 score, and ROC of BayesMulti and ERM at different hardware non-idealities (usability values) on the species prediction dataset. Each method was run 10 times, and the mean (dot) and standard deviation (shaded areas) of accuracy under different usability values are recorded and demonstrated in the curve charts.}
    \label{fig: biology}
\end{figure}

We employed three standard metrics for evaluation: accuracy, Area under the Receiver Operating Characteristic Curve (AUC), and Matthew Correlation Coefficient (MCC). The findings from the HIV dataset (Figure \ref{fig: biology}b) reveal that BayesMulti significantly enhances neutralization predictions for novel HIV antibodies compared to ERM. Notably, across the entire usability spectrum derived from perovskite memristors, BayesMulti's performance remains stable and high across all three metrics. ERM, however, suffers a performance decline at the usability of 0.5 for accuracy, 0.9 for AUC and 0.6 for MCC. Notably, performance degradation for BayesMulti is negligible across the usability range (i.e. 0.93-0.12), and can only be observed at usability $\textless 0.1$. 
Similar trends were observed in the CoVAbDab SARS-CoV-2 dataset (Figure \ref{fig: biology}c), with both methods showing greater sensitivity to hardware non-idealities, indicated by lower overall performance and earlier onset of performance degradation. Nevertheless, BayesMulti substantially outperforms ERM, maintaining a high AUC of 0.8 up to usability $=0.2$, whereas ERM shows a sharp decrease starting at usability $=0.9$. These outcomes highlight BayesMulti's superior effectiveness and robustness in predicting the neutralizing capabilities of previously unidentified antibodies.

We then conducted another biological task involving the prediction of diverse properties and functionalities of glycans. Glycans, complex carbohydrates, play pivotal roles in a multitude of biological processes. Glycans offer a crucial understanding of the physical characteristics and environmental conditions of the organisms they are connected to\citep{lowe2003genetic}. These insights can be extracted by leveraging glycan representations, facilitating the discrimination of distinct taxonomic clusters among organisms. In this context, we employed SweetNet\citep{burkholz2021using}, a graph convolutional neural network, to tackle a species prediction task. Specifically, this task involves predicting the taxonomic kingdom of a glycan based on its representations. Further details regarding the network architecture and noise injection can be found in Figure \ref{fig: SweetNet} and Supplementary Note 5.

We compared BayesMulti with ERM across varying levels of usability, using accuracy, F1 score, and Receiver Operating Characteristic (ROC) as metrics, as depicted in Figure \ref{fig: biology}f. At high usability levels (usability $\textgreater$ 0.8), both BayesMulti and ERM demonstrate high accuracy, exceeding 0.9. However, as usability decreases, ERM's accuracy sharply declines, whereas BayesMulti maintains relatively high values. Notably, when usability reaches 0.4, ERM's predictive capability significantly diminishes (equal to random guessing), with BayesMulti still achieving a robust accuracy of 0.6. This pattern is consistent in other evaluative metrics. Additionally, a t-distributed stochastic neighbor embedding (t-SNE) analysis at the usability of 0.5 visualizes the species classification performance of both methods (Figure \ref{fig: biology}e). BayesMulti distinctly separates glycans into taxonomic kingdoms, unlike ERM, which fails to show discernible clustering by the kingdom. BayesMulti thus demonstrates superior noise resistance, ensuring effective species prediction even with substantial neural network perturbation.

\begin{figure}
    \centering
    \includegraphics[width=\linewidth]{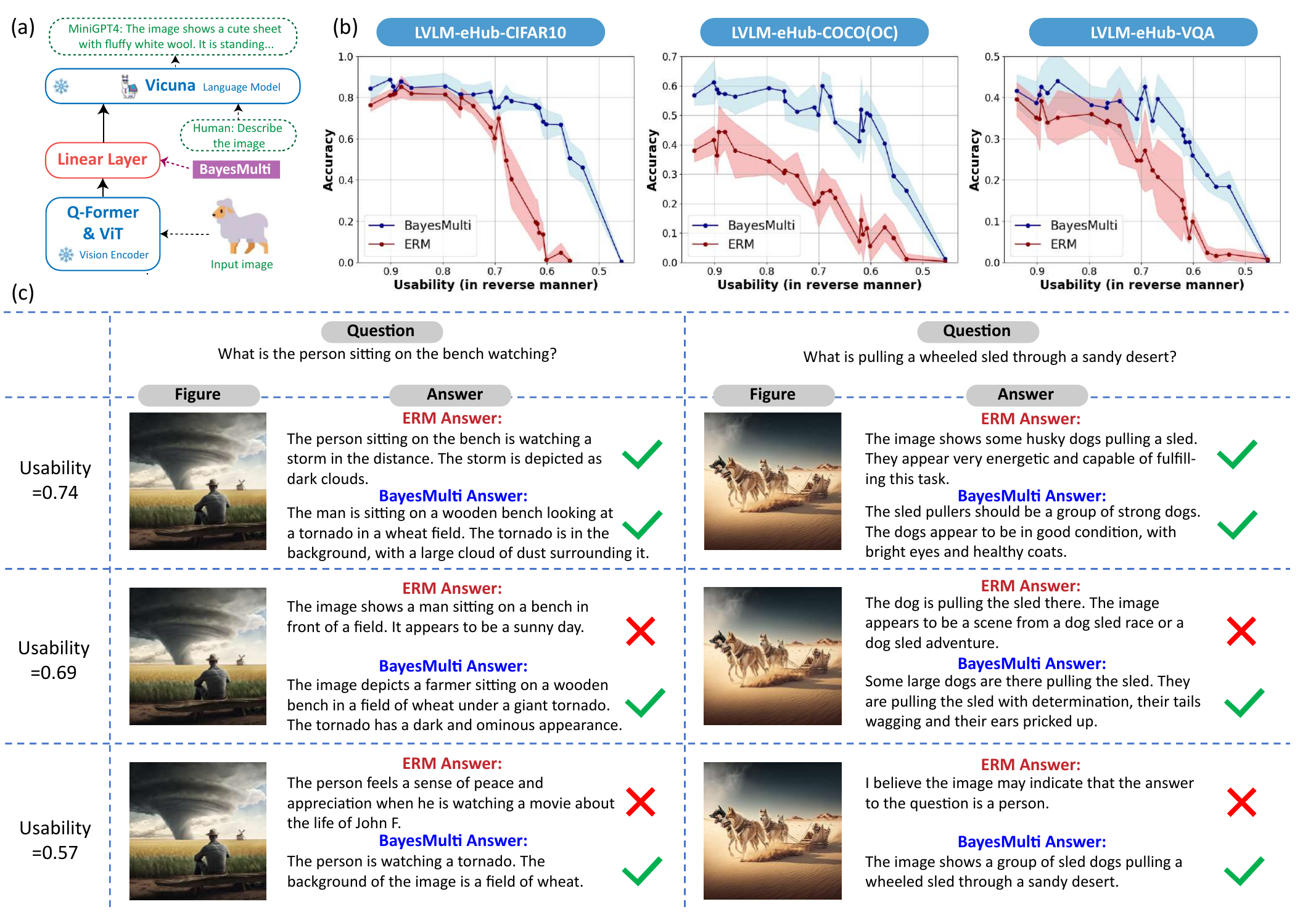}
    \caption{(a) A schematic demonstration of the structure of miniGPT-4, with BayesMulti applied to the linear layers. (b) The performance of BayesMulti and ERM at different hardware non-idealities ($\sigma$ values) on a CIFAR-10 classification task, an MSCOCO object counting task, and an MSCOCO visual question answering task. Each method was run 30 times, and the mean (dot) and standard deviation (shaded areas) of accuracy under different usability levels are recorded and demonstrated in the curve charts. (c) The performance of practical dialogue task of ERM-equipped miniGPT-4 and BayesMulti-equipped miniGPT-4 under different hardware non-idealities (i.e. usability=0.74, 0.69, 0.57). ERM fails to generate correct image descriptions from 0.69, while BayesMulti yields coherent answers aligned with the visual content under all three cases.}
    \label{fig: miniGPT4}
\end{figure}

\subsubsection*{Evaluations on MiniGPT-4}
We finally evaluated the performance of our method on MiniGPT-4\citep{zhu2023minigpt}. As a Large Vision-Language Model (LVLM), it integrates a Vision Transformer (ViT) with a Q-Former and an LLaMA language model, connected via a linear layer. The model undergoes a two-stage training regime: pre-training with a large annotated dataset to learn vision-language interactions—requiring substantial computational power—and fine-tuning with a more refined dataset to reach human-like dialogue precision, which is less computationally demanding and achievable on a single NVIDIA A100 GPU. During training, the vision and language models remain static, and only the intermediate linear layer is adjusted. In our study, we adopt MiniGPT-4's standard structure and implement exclusively the linear layer on the perovskite memristor due to the static nature of the vision and language models (Figure \ref{fig: miniGPT4}a). Similarly, BayesMulti is employed to inject multinomial noise into this layer, thereby enhancing robustness. We then evaluated the model across various hardware noise levels.

To quantitatively evaluate the effectiveness of BayesMulti on MiniGPT-4, we engaged the tiny LVLM-eHub's evaluation framework—a condensed version of LVLM-eHub designed for multimodal task assessment\citep{xu2023lvlm}. Using its metrics, we evaluated MiniGPT-4's performance across classification tasks with the CIFAR-10 dataset, and object counting (OC) and Visual Question Answering (VQA) tasks utilizing the MSCOCO dataset. We complemented these evaluations by conducting subjective assessments of dialogue performance with human evaluators. The outcomes of these comprehensive tests are compiled in Figure \ref{fig: miniGPT4}b, c.

Figure \ref{fig: miniGPT4}b demonstrates that BayesMulti consistently outperforms ERM in tasks involving classification, OC, and VQA. Notably, as usability decreases, ERM's effectiveness markedly diminishes, while BayesMulti-equipped models maintain stable and high scores. This advantage is most pronounced in the OC task, where BayesMulti's accuracy surpasses that of ERM by over 100\% on average across all levels of usability.
In practical dialogue scenarios, MiniGPT-4 is tasked with generating descriptions from images and corresponding questions. Figure \ref{fig: miniGPT4}c reveals that under decreasing usability, the ERM model begins to falter, yielding inaccurate or incoherent responses. Conversely, BayesMulti demonstrates resilience, providing coherent answers aligned with the visual content even amidst comparable usability conditions.
To our knowledge, this represents the first instance of utilizing analog computing for processing large vision language models. Our experiment demonstrates the effectiveness and adaptability of BayesMulti in managing this challenging task, highlighting the potential for implementing LVLMs on analog computing platforms.

\begin{figure}[H]
    \centering
    \includegraphics[width=\linewidth]{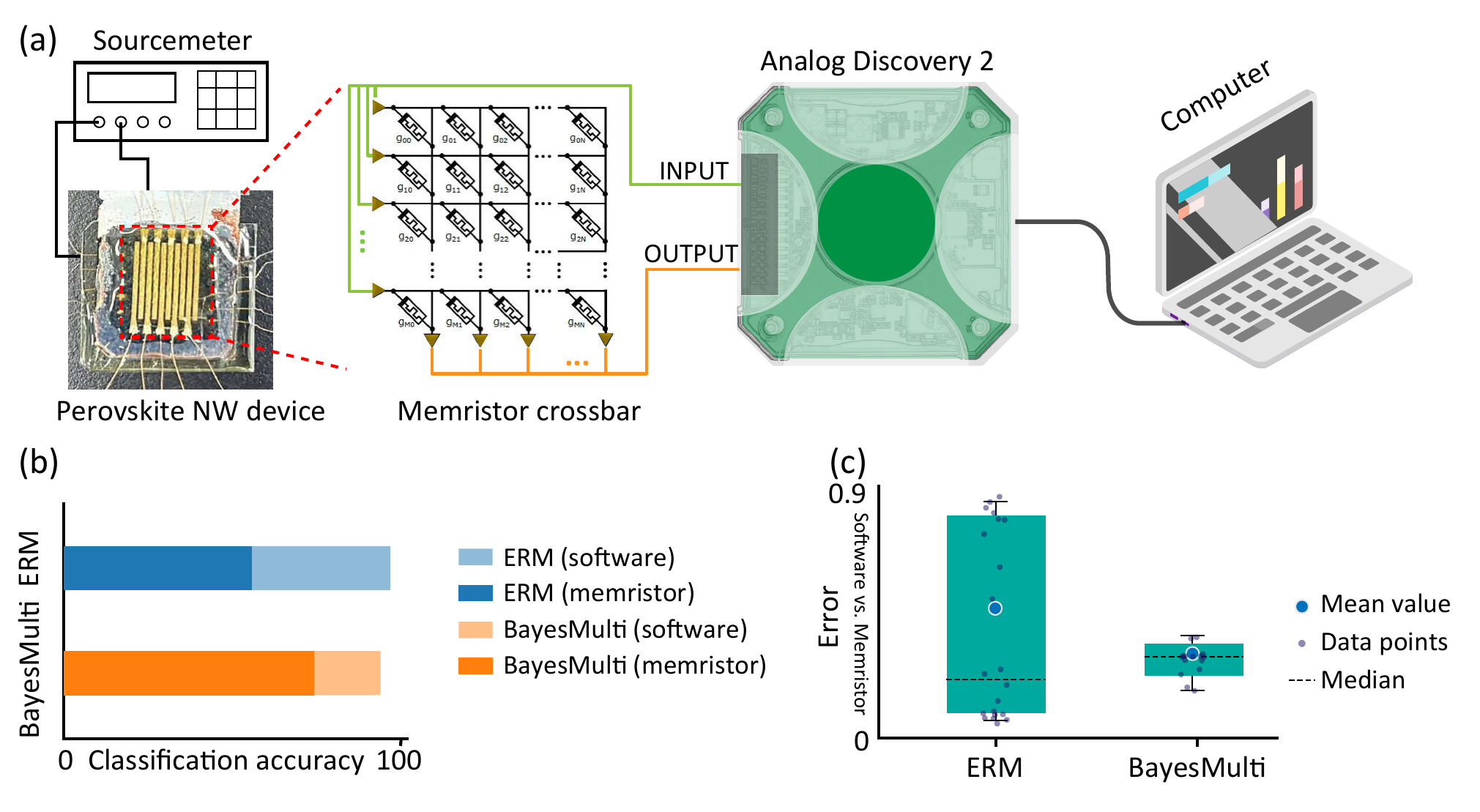}
    \caption{(a) Schematic of the memristor-based analog computing setup, featuring a 10 $\times$ 10 perovskite memristor crossbar designed for multiplication in analog DNN processes. The Analog Discovery 2, serving as both waveform generator and oscilloscope, facilitated analog-digital conversion. It was interfaced with a computer via USB for programmable control over the device's write and read functions during inference.(b) Comparative classification results of ERM and BayesMulti methodologies on a moon-shaped dataset, executed in both software and hardware configurations.(c) Discrepancies in accuracy between software and hardware implementations for ERM and BayesMulti were assessed on the same computational task.}
    \label{fig: circuit_results}
\end{figure}

\subsubsection*{Evaluations on Optimized Perovskite Memristor Crossbar Circuitry }
We finally evaluated the effectiveness of our synergistic development protocol on a real memristor for analog computing. Specifically, a 10 $\times$ 10 memristor crossbar was fabricated based on the optimized material and fabrication conditions, to replace the software multiplication operation in the DNN inference process. The circuit configuration is depicted in Figure~\ref{fig: circuit_results}(a), employing the Analog Discovery 2 for analog-digital conversion. This setup included a perovskite memristor array integrated with peripheral circuitry, enabling programmable write and read operations. Our evaluation focused on a moon-shaped dataset classification task\citep{saito2018maximum} using a network with a single hidden layer. Figure~\ref{fig: circuit_results}(b) presents 20 data points of classification results. Here, light-colored bars represent the classification accuracy computed purely by \emph{NumPy} (software), while dark bars show accuracy attained using the memristor crossbar circuitry (hardware) for multiplication. As shown from the results, although the ERM method delivers a higher theoretical accuracy in software, the accuracy degrades by about 45\% when implementing multiplication operations were implemented in hardware, while our method shows only about 15\% performance degradation. Furthermore, figure \ref{fig: circuit_results}(c) illustrates the discrepancies between hardware and software outputs. Notably, the BayesMulti approach demonstrates lower computational errors and a more concentrated error distribution, underscoring the enhanced stability and reliability of our protocol compared to ERM. Finally, We evaluated the energy efficiency of our perovskite memristor-based analog computing system, using Yao et al.'s methodology\citep{yao2020fully}. Relative to a NVIDIA Tesla V100 GPU, our memristor crossbar (array only) achieves over 270 times higher energy efficiency (27,548 $GOP s^{-1} W^{-1}$). Note that the ADC conversion and the peripheral circuit were not considered in this comparison and they should be further optimized and taken into account in future works.


\section*{Conclusion}
In conclusion, our study presents a unified approach that synergistically enhances the robustness of analog computing, moving beyond the traditional separation of memristor fabrication and application. By employing Bayesian optimization, we have effectively determined optimal fabrication conditions for perovskite memristors with minimized device non-idealities. Concurrently, we introduced BayesMulti, a novel algorithm training strategy that utilizes BO-guided noise injection to improve the robustness of analog DNNs against these device imperfections. Theoretical proofs validate BayesMulti's fault tolerance, and extensive experiments confirm the generalizability and effectiveness of our approach across various deep-learning models and tasks. These include image recognition, sentiment triplet extraction, autonomous driving, biological matching tasks, and even complex LVLMs like Mini-GPT4. This study marks a significant leap in analog computing, showcasing the synergistic integration of device manufacturing and algorithm development to achieve enhanced performance and reliability. Notably, a 10x10 memristor crossbar, fabricated with BO-optimized parameters and trained using BayesMulti, achieved high classification accuracy and  outperformed digital methods in power efficiency by 270 times. Our methodology offers both empirical and theoretical benefits and has broad applicability to different memristor-based analog computing systems and deep-learning algorithms.

\section*{Materials and Methods}

\subsection*{1. Device Fabrication}
\subsubsection*{PAM template fabrication} 

To create Porous Anodic Alumina (PAM) for perovskite nanowire growth, we employed an anodic anodization method as previously described \citep{gu20163d, xu2015novel, ruiz2021revisiting, poddar2021down}. The process began with cutting 0.25 mm thick Aluminum (Al) foils into 20mm × 30 mm chips. These chips were then flattened and sequentially cleaned with acetone and isopropyl alcohol for 10 minutes. The chips underwent electro-polishing in an acidic mixture (comprising 25\% $HClO_4$ and 75\% $CH_{3}CH_{2}OH$ by volume) for 3 minutes to enhance surface cleanliness. The subsequent anodization process varied based on the required pore diameter of the nanowire/quantum wire. \\
\textbf{Large diameter nanowire} The fabrication of PAM templates with 200-nm pore diameters involved anodizing the cleaned Al substrates in a solution (deionized water:ethyleneglycol:$H_3PO_4$ = 200:100:1, by volume) under a 200 V d.c.bias, $10^\circ C$ overnight to promote natural hexagonal ordering. The first anodization layer was then removed using a phosphoric acid mixture (6 wt\% $H_3PO_4$ and 1.8 wt\% $CrO_3$) at $98^\circ C$ for 10 min. A second anodization under identical conditions for varying durations allowed the production of PAM templates with different thicknesses. The length of the final nanowires closely correlates with the anodization time, enabling us to fabricate nanowires ranging from 500 nm to 6 µm. To increase the pore diameter, the template was further etched in a 5\% $H_3PO_4$ aqueous solution at $52^\circ C$. The etching time directly affects the final diameter: 5 minutes for 250 nm and 10 minutes for 300 nm pore diameters.\\
\textbf{Medium diameter nanowire}  PAM templates with pore diameters ranging from 50 to 200 nm were produced using a similar two-stage anodization process in a mixed acid solution. The initial anodization step occurred at $1^\circ C$ for 6 hours in a 1\% $H_3PO_4$ solution with additions of 0.3, 0.1, and 0.03M $C_2H_2O_4$, corresponding to potential values of 120, 150, and 200V. This yielded pore diameters of 150, 100, and 50 nm, respectively. The samples were then immersed in a solution (6\% $H_3PO_4$ and 1.8\% $CrO_3$) at $98^\circ C$ for 10 minutes to remove the first alumina layer. The subsequent anodization followed the same parameters as the first step but varied in duration to produce nanowires of different lengths.\\
\textbf{Quantum wire} Quantum wire (pore diameter: 5-10 nm) fabrication was conducted in a 5 vol \% $H_2SO_4$ solution at a voltage of 10 V. A same two-stage anodization process was carried out and the duration of the second anodization was adjusted to produce quantum wires of different lengths.

\subsubsection*{Barrier thinning and Pb electrodeposition} 
To facilitate the electrochemical deposition of Pb nanoclusters, a voltage ramping down process was conducted to the anodized Al samples to electrochemically thin the residual $Al_2O_3$ barrier layer to approximately 4 nm. This step, essentially another anodization, was performed in the same acid solution used earlier. Using large-pore diameter nanowires as an example, the PAM substrates were immersed in a solution (i.e. same as the anodization solution) at room temperature, controlled by a computer program and a Keithley 2400. Initially, the Keithley 2400 was set to voltage-source mode, progressively increasing the anodization voltage to the set level which is same as the anodization voltage (e.g. 200V for 200-nm diameter nanowires) and recording the initial current $I_0$. It was then switched to current-source mode, reducing the current to $I_0 / 2$, and a gradual decline in current was observed. When the current reduction rate fell below $3 V / min$, the current was further reduced to $I_0 / 4$ to accelerate the voltage decrease. The process ended when the voltage reached 4V, typically within 20-30 minutes.

After barrier thinning, Pb was deposited at the bottom of PAM channels in a three-electrode system with an alternating current method by using a potentiostat. The electrolyte, prepared by dissolving 1.7g of lead(II) chloride ($PbCl_2$) and 25g of trisodium citrate ($Na_3C_6H_5O_7$) in 100mL of water with vigorous stirring, facilitated this process. A 50 Hz sinusoidal voltage with an amplitude ranging from 0.1 to 4 V was applied to maintain a peak current of 6mA during the negative deposition cycle. Post-deposition, the Al substrates were thoroughly rinsed with deionized water to remove residual chemicals. The duration of this process dictated the final thickness of the Pb deposit. 

\subsubsection*{Perovskite nanowire growth and electrode deposition} 
The Pb electrodeposited free-standing PAM substrates were then put into a chemical vapour deposition two-zone tube furnace to react with precursor powder to form perovskite NWs by a Vapor-Solid-Solid-Reaction (VSSR) process as reported before\citep{gu20163d, zhang2021three}. For organic-inorganic perovskite NWs growth (take $\text{MAPbI}_3$ as the example), the procedure involved placing MAI powder at the bottom of a glass bottle and positioning the PAM/Pb substrate at the opening. An adjacent identical bottle was set up with openings touching, creating an enclosed space to maintain high MAI vapor pressure. Both furnace zones were heated to $180^\circ C$ for a 40-minute growth period, with the overall process lasting 3 hours under atmospheric pressure and a continuous argon flow of 150 sccm. For inorganic perovskite NWs growth (take $\text{CsPbI}_3$ as the example\citep{long2023neuromorphic, waleed2017all}), a source powder of CsI and $PbI_2$ with a 3:1 molar ratio was pre-annealed at 450 °C in air for 1 hour. 10g of this mixture was placed in a boat in one furnace zone at $430^\circ C$, while the PAM/Pb substrate was positioned downstream in the second zone at $380^\circ C$. Source vapor was transported by a 20 sccm flow of high-purity Ar gas. The VSSR process occurred in a vacuum environment at 0.55 torr, lasting 3 hours. Upon successful fabrication of perovskite NWs, the top surface of $\text{CsPbI}_3$ was cleaned by ion milling at 400 V with $45^\circ$ angle for 1 hr time. Finally, 200 nm-thick Ag was deposited on top of the perovskite NWs by thermal evaporation process at a pressure of $5 \times 10^{-4}$ Pa and a fast deposition rate of 15 $\AA$ per second to ensure uniformity. A mask was adopted during thermal evaporation to generate individual pixels with an effective cell area of 3.14 $mm^2$.

\subsubsection*{Perovskite NW-based memristors} 
Non-freestanding devices were fabricated on Al substrates using 1.5 mm diameter round shadow masks for Ag evaporation. Ag layers, 50-600 nm thick, were thermally evaporated at a pressure of 5 × $10^{-4}$ Pa and a rate of 15 Å/s to ensure uniformity. For all I-V measurements, 5X5 arrays with Al as the co-counter electrode were utilized.

For the freestanding 10×10 array devices, an additional layer of approximately 4 nm $Al_2O_3$ was deposited on top of the barrier layer using atomic layer deposition before Au counter electrode evaporation. This involved 1 mm wide finger electrodes for Au deposition under similar conditions. Post flip-over, wire-bonding, and epoxy curing as previously described, Ag deposition was conducted using similar finger electrodes\citep{gu20163d}. The effective cell area of the freestanding sample was defined by the intersection of 1 mm long Ag and Au electrodes.
 
\subsection*{2. Device Characterization}
\textbf{SEM imaging} The cross-sectional images of the PAM and perovskite NWs samples were collected by using a field emission scanning electron microscope ZEM15-Desktop in back-scattered electron (BSE) mode.\\
\textbf{Electrical measurements} The cyclic I-V characteristics were measured by Keithley 6487 with home-built LABVIEW programs.

\subsection*{3. Simulations of device non-idealities}
\subsubsection*{Simulating Non-monotonic Non-ideality} 
In analog computing, neural networks weight ($\theta$) are represented by memristors' conductance ($C$) which is modulated by the input trains of pulses. The non-idealities between conductance $C$ and the number of charging pulses include non-linearity and non-monotonicity. Non-linearity can be compensated by designing the mapping scheme between $\theta$ to $C$ or changing pulse shapes\citep{joksas2022nonideality, xi2020memory, st2014general}. 
However, non-monotonicity limits the working region of the memristor, thus constraining the memristor's multi-level resistance characteristic and analog computation precision. To quantify the level of non-monotonicity, we first extract the length of the Longest Conductance Increasing Subsequence---LCIS from the conductance curve, which represents the operative segment of the memristor. Experimentally, the memristor is initially charged to reach the starting point of LCIS, serving as the reference for the desired weight. Incremental charging is then applied to enhance the relative conductance until the target weight is achieved. Notably, an excessively short LCIS demands exceptionally precise charge control, posing challenges in circuit design. Therefore, we define a minimum operative conductance length ($Require\_len$).
If the length of LCIS, $l_{LCIS} \geq Require\_len$, non-monotonic noise is absent; however, if $l_{LCIS} < Require\_len$, it is present and leads to weight drifting in analog DNNs. In such cases, we apply the mapping $\theta_{mono} \Leftarrow (C-C_{min})/(C_{max}-C_{min}) \times \vert \theta \vert_{\max}$, where $\theta_{mono}$ represents the true weight of the analog DNN, $C$ is the conductance exceeding the operative segment, $C_{max}$ is the maximum conductance, $C_{min}$ is the minimum conductance, $\vert \theta \vert_{\max}$ is the maximum absolute value in a $\theta$ matrix for a matrix-vector product in analog computing. Note that this is not the only way considering the non-monotonic ideality but requires a few hardware modifications. We define the non-monotonic non-ideality factor as: 
\begin{equation}
\frac{l_{L C I S}}{Require \textunderscore len}
\end{equation}

\subsubsection*{Simulating Stochastic Non-ideality} 
In practice, memristors often deviate from ideal I-V characteristics, exhibiting unavoidable cycle-to-cycle variations in each I-V measurement. We attribute these variations to stochastic non-ideality, assuming it follows a log-normal distribution:
\begin{equation}
\theta^{\prime} \leftarrow \theta e^\lambda, \quad \lambda \sim \mathcal{N}\left(0, \sigma^2\right)
\end{equation}
where $\theta^{\prime}$ denotes the weights of the analog DNN after memristance drift, following a log-normal distribution \citep{chen2017accelerator, LeeVLSIT}. 
$\theta$ represents the expected weights of the neural network. $\sigma$ denotes the intensity of the stochastic noise, which needs to be estimated from the measured outcomes. Note that $\theta^{\prime} / \theta = e^\lambda$, therefore, we only need to consider the ratio between the drifted weight and the expected weight to estimate $\sigma$. Given that weight is denoted by conductance in analog DNNs, we can employ Maximum Likelihood Estimation (MLE) to ascertain the ratio of memristor conductance values to the mean conductance for accurate $\sigma$ estimation from different cycles' measurements.
Specifically, we assume that the ratio between the conductance value in cycle $i$, $c_i$, and the mean value, $\bar{c}$, follows a log-normal distribution: 
$$
\frac{c_i}{\bar{c}}\sim\textrm{log-normal}(0, v),
$$
where $v$ is the variance from the maximum likelihood estimation according to the I-V measurement data across different cycles. Further details of the calculation are elaborated in the Supplementary Note 1.

\subsubsection*{Comprehensive Indicator for Perovskite Memristors' Non-idealities}
Finally, we introduce the concept of ``usability" to assess a typical memristor's suitability for analog computing. This metric is defined mathematically as:
\begin{equation}
\text { Usability }=\frac{l_{L C I S}}{Require \textunderscore len} \cdot \exp(-\sigma)
\end{equation}
It incorporates both the non-monotonic non-ideality factor, represented by $\frac{l_{L C I S}}{Require \textunderscore len}$, and the stochastic non-ideality factor, denoted by $\sigma$. ``Usability" is employed as the optimization objective for the subsequent BO process, serving as an indicator for the prosperity of a specific perovskite manufacturing design.

\section*{Acknowledgments}
\textbf{Funding} QG is grateful for the support from Shanghai Artificial Intelligence Laboratory and the National Key R\&D Program of China (Grant NO.2022ZD0160100). NY acknowledges the funding from the National Science Foundation of China (Grant No. 62106139). 
\textbf{Author contributions} QG and QS contributed equally to this work. QG fabricated the memristors and measured their characteristics. QG and QS conducted the noise-injecting experiments, and task evaluations, as well as analyzed data and prepared figures. YW and LY conducted task evaluations. NY conceived the concept, supervised the work, and provided the theoretical analysis. JZ helped to conduct theoretical proofs. QS, YW and LY conducted this work during their internship at the Shanghai Artificial Intelligence Laboratory. All the authors contributed to results analysis and manuscript writing.

\section*{Competing interests} 
The authors declare no competing interests.

\section*{Data and material availability} All data needed to evaluate the conclusions in the paper are present in the main text or the Supplementary Information. Training algorithm implementation code will be released upon publication.

\backmatter
\newpage


\newpage
\begin{appendices}
\textbf{\title{\begin{center}{\large Supplementary Information}\end{center}}}

\section*{Supplementary Note 1: Detailed Explanation of Hardware Non-idealities}
In this paper, hardware non-idealities are simulated by measuring real resistance fluctuations, emulating the phenomenon of weight drifting observed in memristors. We focus on the positive quadrant of the I-V curve for clarity. As shown in Figure~\ref{fig:hwnoise_illustrate}, the hardware non-idealities mainly arise from two sources. The first is the variations of conductance across different cycles of charging, denoted by the shaded area. In practical circuits, while an ``average memristor'' charging scheme is typically employed, deviations from this norm can lead to discrepancies between targeted and actual conductance. This variance is primarily due to stochastic thermal noises, measurement errors, etc. The second is the non-monotonic behavior in the curve, where resistance unexpectedly decreases with positive charging currents, causing programming errors that can lead to significant weight reduction. Consequently, only the monotonic section of the curve is viable for analog computing. The extent of this monotonic part is crucial; it must be sufficiently long to prevent premature saturation and ensure feasibility in analog computing circuit designs. This second type of non-ideality is generally attributed to imperfections in memristor fabrication.

In our experiments, we showcase examples of actual measurement outcomes. Figure~\ref{fig:good_device} illustrates a well-performing perovskite NW-based memristor, optimized using our Bayesian optimization approach. Its conductance characteristics, depicted in Figure~\ref{fig:mean_deviation_good}, reveal a consistent monotonic increase in conductance with minimal variation across different cycles. Conversely, Figures \ref{fig:0.7um_device} and \ref{fig:0.7um mean_deviation_good} present data from an unoptimized perovskite NW-based memristor. Here, the conductance trend is non-monotonic and exhibits significant variance between cycles, indicating considerable stochastic noise.

\subsection*{Statistical Modeling of Hardware Non-idealities}
In this section, we first discuss how to simulate the non-monotonic non-ideality and then the stochastic non-ideality. Finally, we will derive an overall numeric indicator for the suitability of any memristor device for analog computing.

\textbf{Simulating non-monotonic non-ideality}
For simplicity and without loss of generality, we defined the minimum operative conductance range $Require\_len$ in the curve. If the length of the longest conductance increasing subsequence (LCIS) exceeds $Require\_len$, then, the non-monotonic non-ideality is absent. Otherwise, we apply the following mapping for points outside of the LCIS region:
\begin{equation}
    \theta_{mono} \Leftarrow (C-C_{min})/(C_{max}-C_{min}) \times \vert \theta \vert_{\max}
\end{equation}
where $\theta_{mono}$ represents the true weight of the analog neural network, $C$ is the conductance exceeding the operative segment according to the target $\theta$, $C_{max}$ is the maximum conductance, $C_{min}$ is the minimum conductance, $\vert \theta \vert_{\max}$ is the maximum absolute value in a weight matrix for a matrix-vector product in analog computing. This maps the non-monotonically increasing $C$ according to its relative position between $C_{max}$ and $C_{min}$. Note that there may exist other ways for considering the non-monotonic non-ideality by introducing complex circuitry design. However, due to the limited budget for circuit complexity for better energy efficiency, we choose this mapping for simplicity.

\textbf{Simulating stochastic non-ideality} 
As shown in Figure~\ref{fig:hwnoise_illustrate}, conductance fluctuations in different cycles and pixels can lead to degenerated performances in analog computing. As discussed in the main text, we model the stochastic non-ideality with a log-normal distribution \citep{chen2017accelerator,LeeVLSIT}:

\begin{equation}
\theta^{\prime} \leftarrow \theta e^\lambda, \quad \lambda \sim \mathcal{N}\left(0, \sigma^2\right)
\end{equation}

where $\theta^{\prime}$ denotes the weights of the neural network after drifting, following a log-normal distribution.
$\theta$ represents the expected weights of the neural network. $\sigma$ denotes the standard deviation of the Gaussian noise. Note that $\log \left(\theta^{\prime}/\theta \right)= \lambda \sim \mathcal{N} \left(0, \sigma^2 \right)$. As $\theta$ has a one-to-one correspondence to the conductance $C$, we only need to measure the variances of $C$, which also follows $\log\left(C^{\prime} / C \right)= \lambda \sim \mathcal{N}\left(0, \sigma^2\right)$, where $C^{\prime}$ is the conductance after drifting, and $C$ is the expected conductance. We derive the maximum likelihood estimation (MLE) of $\sigma$ and its upper 95 percent confidence bound as follows:
\begin{equation}
    \sigma_{\text{MLE}}^2 = \frac{1}{n} \sum_{i=1}^{n}  [\log \left(\frac{C^{\prime}_i}{C_i} \right)]^2
\end{equation}
where $C^{\prime}_i$ is the $i$-th measured drifted conductance, $C_i$ is the corresponding $i$-th average conductance, and $n$ is the total number of measurements. As $\sigma_{\text{MLE}}^2$ follows a $\chi^2$ distribution, the upper 95 percent confidence bound is:
\begin{equation}
  \sigma_{95}^2 = \frac{(n-1)\sigma_{\text{MLE}}^2}{\chi^2_{97.5}}
\end{equation}
where $\chi^2_{97.5}$ is the normalization constant for 95 percent upper confidence bound. In the experiments, we use $\sigma_{95}$ instead of $\sigma_{\text{MLE}}$ to have a more conservative estimation of variance. This further ensures the robustness of our experiment results.


\section*{Supplementary Note 2: Process of Bayesian Fabrication Optimization}

\subsection*{Search Space Definition}
In our research, Bayesian optimization was employed to identify the optimal design parameters for perovskite NW-based memristors used in analog computing. Based on previous studies and our expertise, we evaluated several physical parameters likely influencing memristor performance, including (1) the morphology of NWs, (2) the thickness of barrier layer, (3) the type of perovskite, (4) the quality of perovskite (e.g. side product, crystalline phase), (5) the compatibility of NWs and perovskite (i.e. overfill/underfill/crystal expansion) and (6) the characterization conditions. We determined that the barrier layer thickness of $Al_2O_3$, being only 4 nm, offers limited scope for variation and is unlikely to significantly impact memristor performance. Consequently, considering the experimental parameters that would affect the above physical conditions, we finally select the following five key variables: (1) perovskite types ($MAPbX_3, X=Cl, Br, I; CsPbI3, FAPbI_3, FAPbBr_3$, (2) NW length ($0.6 \mu m-3\mu m$), (3) NW diameter (10 nm-300 nm), (4) lead electrodeposition (Pb ED) time (5 min-30 min, and (5) Ag thickness (50 nm-600 nm). Though most factors are continuous variables, considering the experimental feasibility and reproductivity, we set several steps among the defined range. For example, ``NW length" consists of 0.6$\mu$m, 1$\mu$m, 1.2$\mu$m, 1.5$\mu$m, 1.8$\mu$m, 2$\mu$m, 2.5$\mu$m and 3$\mu$m. These variables define our final search space, comprising 8400 potential experimental configurations.

\subsection*{The Process of Bayesian Optimization}
The primary objective of Bayesian optimization can be formulated as:

\begin{equation}\label{equ:objective}
    x^* = {\arg\max}_{x \in A} f(x),
\end{equation}
In this formulation, $x$ represents a $d$-dimensional vector within a feasible domain $A$, and $x^{*}$ denotes the optimal solution. The function $f$, key to this optimization process, is characterized by its flexibility; it is defined solely by the input-output pairs it maps, without imposing structural constraints. This lack of constraints implies an absence of prior knowledge about the relationship between inputs and outputs, making traditional gradient-based optimization methods inapplicable.
Bayesian optimization addresses this challenge by employing a posteriori knowledge, making it a powerful and novel optimizer for automating machine learning tasks.

Figure \ref{fig:BO_process}(a) showcases the Bayesian optimization algorithm's step-by-step process. It involves fitting data $D$ to produce the sample function $u$, where each point in the search space is linked to an input $x$. The output from this function is a predictive probability distribution over the target objective function, formulated through Gaussian process regression. This model encapsulates the objective function’s distribution using the accumulated data from the optimization process. The expression $g_{D}(x_{})$ offers a probabilistic depiction of the objective function $f_{}$, informed by the data at hand, and from this depiction, the sample function $u$ is derived:

\begin{equation}
    g_{D}(x_{*}) \sim N(\mu(x_*), \sigma^2(x_*))
\end{equation}

Here, $\mu(x_*)$ represents the mean as predicted by the Gaussian Process model, and $\sigma^2(x_*)$ denotes the variance. The notation $N(.)$ refers to the Gaussian distribution. The following formulas will expound on the development of a Gaussian Process model.

\paragraph{General Knowledge}

A Gaussian Process (GP) consists of a collection of random variables, each adhering to a Gaussian distribution. The distribution for the entire set of random variables in the GP is characterized by its mean, $\mu(x) = E[f(x)]$, and covariance:

\begin{equation}
    K(x,x_*) = E[(f(x) - \mu(x))(f(x_*) - \mu(x_*))]
\end{equation}

which is instrumental in defining the relationship between input variables and the output probability. The surrogate GP model is trained using pairs of variable sets and their corresponding metric scores derived from the objective function. In this setup, the variable set acts as the input, and the metric score serves as the output. This training enables the surrogate model to approximate the objective function's behavior effectively.

\paragraph{Mathematical Inference}
The covariance function, often a squared exponential or Gaussian Kernel, is pivotal in constructing a practical Gaussian Process Model (GPM). It models the covariance between two points in the input space as a function of their distance:

\begin{equation}
    K(x,x_*) = exp(-\frac{1}{2\sigma^2}\Vert{x - x_*}\Vert^2)
\end{equation}

This function quantifies the similarity between two samples, establishing a prior-based link between iterations. Integrating the above equations, the posterior distribution after $n$ iterations is $ f_n \mid x_n, D \sim N(\mu(x_n), \sigma^2(x_n))$, where:

\begin{equation}
    k_* = [k(x_*,x_1) \;...\; k(x_1,x_*)]^T
\end{equation}

\begin{equation}
    \mu(x_n) = \mu(x_1) + k_{f_n}^T(K_f + \sigma^2_{noise}I)^{-1}\cdot(y - \mu(x_1)\cdot1)
\end{equation}

\begin{equation}
    \sigma^2(x_n) = k_f(x_n,x) - k_{f_n}^T(K_f + \sigma^2_{noise}I)^{-1}k_{f_n}
\end{equation}

Here, $K_f$ is the Gram matrix\citep{garnett2023bayesian} with elements from the covariance function $k_f$.

These formulas complete the process of data fitting to derive the sampling function $u(x,g_D)$, leading to the update of $x$ as per $argmax_x(u(x,g_D))$. This aims to align the new sampling points closely with the extreme value points obtained by model fitting. In Bayesian optimization, these utility functions, also known as acquisition functions, are crucial. The adoption of a GPM, informed by prior knowledge, enables this method to surpass traditional techniques by ensuring quicker convergence with fewer iterations.

In essence, Bayesian optimization uses a GPM that iteratively refines the prior by assimilating previously sampled data. By harnessing this historical information, it efficiently identifies new sampling points at each iteration, streamlining the optimization process.


\section*{Supplementary Note 3: Theoretical analysis}
In this section, we will prove the robustness of the proposed algorithmic scheme against weight drifting. We first introduce the proposed randomized version of the analog DNN model: $f_{\pi_{0}}(\theta_{0}, x) := \Ebb_{\eta \sim \pi_{0}}[f(\theta_{0}*\eta, x)]$, where $x$ is the input data; $\theta_{0}$ is the original unperturbed parameters of the analog DNN; $\pi_{0}$ is the random noise injected in the analog DNN, \textit{i.e.}, multinomial distribution; $*$ is the multiplication for applying the random noise. In the following parts, we write $f_{\pi_{0}}(\theta_{0}, x)$ as $f_{\pi_{0}}(\theta_{0})$ for brevity. For simplicity and without loss of generality, we consider the case of a 0-1 classification problem. We want to prove that if $f_{\pi_{0}}>\frac{1}{2}$, then for any perturbation on the analog DNN's parameter $\delta$ within some ranges, the prediction of the analog DNN remains the same (i.e. $f_{\pi_{0}}(\theta_{0} * \delta) > \frac{1}{2}$).

To derive a tractable lower bound for $f_{\pi_{0}}(\theta_{0} * \delta)$ and deduce $\delta$, we relax $f$ in the functional space $\Fcal=\{\hat{f}:\hat{f}(\theta) \in [0,1]\}$ which is the set of all functions bounded in $[0,1]$, along with a equality constraint at the original function $f$:
\begin{align}
    &\min_{\deltaB \in \Bcal} f_{\pi_{0}}(\theta_{0} * \delta) \geq \min_{\hat{f} \in \Fcal} \left\{\min_{\delta \in \Bcal} \hat{f}_{\pi_{0}}(\theta_{0} * \delta)\right\} \label{ineq:constrained} \\
    & ~\mathrm{s.t.} ~~ \hat{f}_{\pi_{0}}(\theta_{0})=f_{\pi_{0}}(\theta_{0}) \nonumber
\end{align}

\begin{theorem}
\label{thm:lagrangian}
\textbf{(Lagrangian)} Denote by $\pi_{\delta}$ the distribution of $\eta * \delta$, solving Inequality~\ref{ineq:constrained} is equivalent to solving the following problem:
\begin{align}
    \Lcal &=\min_{\hat{f}\in\Fcal}\min_{\delta \in \Bcal} \max_{\lambda \in \Rbb}\left\{\hat{f}_{\pi_{0}}(\theta_{0} * \delta)-\lambda(\hat{f}_{\pi_{0}}(\theta_{0})-f_{\pi_{0}}(\theta_{0}))\right\} \notag \\
     & \geq \max_{\lambda \geq 0} \left\{\lambda f_{\pi_{0}}(\theta_{0})-\max_{\delta \in \Bcal}\mathbb{D}_{\mathcal{F}}(\lambda\pi_{0},\pi_{\delta})\right\}
     \label{eq:lagrangian}
\end{align}
where $\mathbb{D}_{\mathcal{F}}(\lambda\pi_{0},\pi_{\delta})$ is:
\begin{align}
    &\mathbb{D}_{\mathcal{F}}(\lambda\pi_{0},\pi_{\delta}) \nonumber \\
    &=\max_{\hat{f} \in \Fcal}\left\{ \lambda\Ebb_{\eta \sim \pi_{0}} [\hat{f}(\theta_{0} * \eta)]-\Ebb_{\eta \sim \pi_{\delta}} [\hat{f}(\theta_{0} * \eta)]\right\}  \\
    &= \sum [\lambda\pi_{0}(\eta)-\pi_{\delta}(\eta)]_{+} 
\end{align}
\end{theorem}

Theorem~\ref{thm:lagrangian} is proved with the Min-Max theorem. Now we are able to analyze the robustness induced by the injected multinomial distribution. For $\eta \sim \pi_0$, $\eta=0$ with probability $p_1$, $\eta=0.5$ with probability $p_2$, and $\eta=1$ with probability $1-p_1-p_2$. 

\begin{lemma}
\label{lemma:Dfbound}
\textbf{($\mathbb{D}_{\mathcal{F}}$ bound of multinomial distribution)}
    Let $\eta$ be multinomial distributed. Let $k$,$l$,$m$ be the number of 0, 0.5 and 1 in $\delta$. We have
    \begin{equation}
    \mathbb{D}_\Fcal(\lambda\pi_0, \pi_\delta) = \lambda(1-p_1^{\Theta-m-l}(1-p_2)^l) + [\lambda-p_1^{-k}]_+p_1^{\Theta-m-l}(1-p_2)^l
    \label{MUL}
    \end{equation}
    where $\Theta$ is the dimension of $\theta_0$.
\end{lemma}

\begin{proof}[Proof of Lemma 3]
    Without loss of generality, assume
    \begin{align}
        \begin{cases}
            \delta_i = 0,\ for\ i=1,\dots,k \\
            \delta_i = 0.5,\ for\ i=k+1,\dots,k+l \\
            \delta_i = 1,\ for\ i=k+l+1,\dots,k+l+m \\
            \delta_i \neq 0,\ 0.5\ or\ 1,\ for\ i=k+l+m+1,\dots,\Theta
        \end{cases}
    \end{align}
    Since we require $\pi_0(\eta)>0$ to have $[\lambda\pi_0(\eta)-\pi_{\delta^*}(\eta)]_+>0$, we could partition the support space $\Hcal=\{\eta\in\{0,0.5,1\}^\Theta\}$ into $\Hcal_1:=\{\eta\in\mathbb{R}^\Theta:\lambda\pi_0(\eta)>0, \pi_\delta(\eta)=0\}$ and $\Hcal_2:=\{\eta\in\mathbb{R}^\Theta:\lambda\pi_0(\eta)>0, \pi_\delta(\eta)>0\}$. \\
    Note that $\Hcal_1=\{\eta\in\{0,0.5,1\}^\Theta:\exists\eta_i>0$ for $i=1,\dots,k$ or $\exists \eta_i=0.5$ for $i=1,\dots,k+l,k+l+m+1,\dots,\Theta$. or $\exists \eta_i=1$ for $i=1,\dots,k,k+l+m+1,\dots,\Theta\}$\\
    And $\Hcal_2=\{\eta\in\{0,0.5, 1\}^\Theta:\eta_i=0\ \forall i=1,\dots,k,k+m+l+1,\dots,\Theta$ and $\eta_i\neq 0.5\ \forall i=k+1,\dots,k+l\}$.

  \footnotesize{
    \begin{align}
        &\mathbb{D}_\Fcal(\lambda\pi_0, \pi_\delta) = \sum_{\eta \in \Hcal_1} \lambda \pi_0(\eta) + \sum_{\eta \in \Hcal_2} [\lambda \pi_0(\eta)-\pi_\delta(\eta)]_+ \\
        &=\lambda(1-p_1^k(1-p_2)^lp_1^{\Theta-k-m-l})+\sum_{a=0}^{l}\sum_{b=0}^{m}\sum_{c=0}^{m-b} \notag \\
        &[\tbinom{l}{a}\tbinom{m}{b}\tbinom{m-b}{c}(\lambda p_1^{\Theta-m-l+a+b}(1-p_1-p_2)^{c+l-a}p_2^{m-b-c}-p_1^{\Theta-m-l+a+b-k}(1-p_1-p_2)^{c+l-a}p_2^{m-b-c})]_+ \notag \\
        &= \lambda(1-p_1^{\Theta-m-l}(1-p_2)^l) + [\lambda-p_1^{-k}]_+\sum_{a=0}^{l}\sum_{b=0}^{m}\sum_{c=0}^{m-b}\tbinom{l}{a}\tbinom{m}{b}\tbinom{m-b}{c}p_1^{\Theta-m-l+a+b}(1-p_1-p_2)^{c+l-a}p_2^{m-b-c} \notag \\
        &= \lambda(1-p_1^{\Theta-m-l}(1-p_2)^l) + [\lambda-p_1^{-k}]_+p_1^{\Theta-m-l}\sum_{a=0}^{l}\sum_{b=0}^{m}\sum_{c=0}^{m-b}\tbinom{l}{a}\tbinom{m}{b}\tbinom{m-b}{c}p_1^{a+b}(1-p_1-p_2)^{c+l-a}p_2^{m-b-c} \notag \\
        &= \lambda(1-p_1^{\Theta-m-l}(1-p_2)^l) + [\lambda-p_1^{-k}]_+p_1^{\Theta-m-l}\sum_{a=0}^{l}\tbinom{l}{a}(1-p_1-p_2)^{l-a}\sum_{b=0}^{m}\tbinom{m}{b}p_1^{a+b}(1-p_1)^{m-b} \notag\\
        &= \lambda(1-p_1^{\Theta-m-l}(1-p_2)^l) + [\lambda-p_1^{-k}]_+p_1^{\Theta-m-l}\sum_{a=0}^{l}\tbinom{l}{a}(1-p_1-p_2)^{l-a}p_1^a \notag\\
        &= \lambda(1-p_1^{\Theta-m-l}(1-p_2)^l) + [\lambda-p_1^{-k}]_+p_1^{\Theta-m-l}(1-p_2)^l
    \end{align}
}
\end{proof}

\textbf{(Robustness analysis)}
Note that $\mathbb{D}_\Fcal$ is decreasing function with respect to $k$, substituting this into Lagrangian:
\begin{align}
    \Lcal &\geq \max_{\lambda\geq 0}\{\lambda f_{\pi_0}(\theta_0)-\max_{\delta\in\Bcal}\mathbb{D}_\Fcal(\lambda\pi_0, \pi_\delta)\} \\
    &= \max_{\lambda\geq 0}\min_{\delta\in\Bcal}\{\lambda(f_{\pi_0}(\theta_0)-1+p_1^{\Theta-m-l}(1-p_2)^l)-[\lambda-1]_+p_1^{\Theta-m-l}(1-p_2)^l\} \label{ieq32}
\end{align}

According to the definition of $\Bcal$, we have $\Theta-m-l \leq r$. Denote the right-hand side of \ref{ieq32} as $RHS$.

Case 1. $\lambda \leq 1$. 
\begin{align}
RHS = \max_{0\leq\lambda\leq1}\min_{\delta\in\Bcal}\lambda(f_{\pi_0}(\theta_0)-1+p_1^{\Theta-m-l}(1-p_2)^l). 
\end{align}

\quad Case 1.1 $\min_{\delta\in\Bcal} f_{\pi_0}(\theta_0)-1+p_1^{\Theta-m-l}(1-p_2)^l \leq 0$. Then $RHS=0<1/2$.

\quad Case 1.2 $\min_{\delta\in\Bcal} f_{\pi_0}(\theta_0)-1+p_1^{\Theta-m-l}(1-p_2)^l > 0$. The minimum is attained at $\lambda=1$:
\begin{align}
RHS = \min_{\delta\in\Bcal}f_{\pi_0}(\theta_0)-1+p_1^{\Theta-m-l}(1-p_2)^l
\end{align}
Since $1>1-p_2>p_1$, the funtion increases with respect to $l$ and $m$. And $\frac{1}{p_1}>\frac{1-p_2}{p_1}$, therefore the function increases feaster w.r.t. to $m$ than $l$. To achieve the lower bound under the constraint that $m+l\geq \Theta-r$, we have $m=0$ and $l=\Theta-r$. Therefore 
\begin{align}
    &RHS=f_{\pi_0}(\theta_0)-1+p_1^r(1-p_2)^{\Theta-r} > 0.5 \\
    &\Leftrightarrow r \leq \frac{\ln(1.5-f_{\pi_0}(\theta_0))-\Theta\ln(1-p_2)}{\ln p_1-\ln (1-p_2)}
\end{align}

Case 2. $\lambda \geq 1$.
\begin{align}
RHS=\max_{\lambda\geq1}\min_{\delta\in\Bcal} \lambda(f_{\pi_0}(\theta_0)-1)+p_1^{\Theta-m-l}(1-p_2)^l=\min_{\delta\in\Bcal}f_{\pi_0}(\theta_0)-1+p_1^{\Theta-m-l}(1-p_2)^l
\end{align}
the situation is same as Case 1.2.

\textbf{(Robustness guarantee)} From the analysis above, we derive the robustness guarantee of the proposed method. From this result, intuitively, we should increase $p_1$ and $p_2$ as much as possible to enlarge the allowable perturbation range of DNN's weights. However, too large $p_1$ and $p_2$ may lead to smaller $f_{\pi_0}(\theta_0)$, constraining the maximum allowable perturbation range. This necessitates the introduction of a Bayesian optimization procedure to search for optimal $p_1$ and $p_2$. Additionally, compared with the Bernoulli distribution, the multinomial distribution can enlarge the range of robustness by introducing additional flexibility in noise injection. However, the additional parameters introduced by the multinomial distribution may slow down the Bayesian optimization processes. Thus, we use the multinomial distribution with three possible values to achieve a balance between performance and optimization time.


\section*{Supplementary Note 4: Implementation details of PerovskiteMemSim}

\lstset{
    basicstyle=\ttfamily,
    keywordstyle=\color{blue}\bfseries,
    commentstyle=\color{gray},
    stringstyle=\color{red},
    numbers=left,
    numberstyle=\tiny\color{gray},
    breaklines=true,
    breakatwhitespace=true,
    showstringspaces=false,
    tabsize=4,
    frame=single,
    captionpos=b
}

\begin{lstlisting}[language=Python, caption=Python code implementation and prototypes for PerovskiteMemSim, label=code:example]
## using case:
'''
f_name: file name of the I-V test data file (*.csv), corresonding to certain memristor farbricated, used for noise analysis
model: certain instance of an implemented artificial neural network, to whom the simulation will be implemented
'''
weight_mapping(f_name, model, device='cuda')



## implementation and prototypes:

from scipy import interpolate
import matplotlib.pyplot as plt
import pandas as pd
import matplotlib.pyplot as plt
import numpy as np
from pykalman import KalmanFilter
import torch
import torch.nn as nn
import torch.nn.functional as F
from scipy.stats import norm

def weight_mapping(f_name, model, noise_sigma, device='cuda'):
    """
    Map target weight to true weight. 
    The difference between the target weight and the true weight is due to two factors:
    1.Random heat noises
    2.Non-monotonic characteristics of conductance-Q curves

    To achieve the above functionality, we first have to measure the conductance-Q curve. 
    As the measurement is affected by noises, we measure the curve for several times and get
    the mean value. 
    """
    c_mean_smooth = calculate_smoothed_cmean(f_name)

    max_index = c_mean_smooth.shape[0]
    c_max = c_mean_smooth.max()
    c_min = c_mean_smooth.min()

    """
    We calculate the LCIS of the curve to find a monotonicly increasing part of the curve.
    """

    start_index, end_index= LCIS(c_mean_smooth) 
    mono_len = end_index - start_index

    required_len = 35

    ratio = np.ones(required_len)
    if mono_len < required_len: # Non-monotonic characteristics 
        increase_indices = np.arange(start_index, end_index, 1)
        end_index = start_index + required_len
        if end_index > max_index:
            start_index = start_index - end_index + max_index
            end_index = max_index

        for i in range(required_len):
            idx = i + start_index
            # non-monotonic characteristics
            ratio[i] = 1 if idx in increase_indices else (c_mean_smooth[idx]-c_min)/(c_max-c_min)

    # noise_sigma = mlesigma(f_name, start_index, end_index)
    gassian_kernel = torch.distributions.Normal(0.0, noise_sigma)
    with torch.no_grad():
        for theta in model.parameters():
            abstheta = torch.abs(theta) 
            normalized_theta = abstheta / (torch.max(abstheta) + 1e-8) 

            theta_index = normalized_theta * (required_len-1)
            theta_index = theta_index.type(torch.LongTensor) 
            noise_index = normalized_theta * 100
            noise_index = noise_index.type(torch.LongTensor)
            noise_index[noise_index >= 100] = 99

            theta_ratio = torch.Tensor(ratio)[theta_index].cuda() 

            mul_ = theta_ratio * torch.exp(
                gassian_kernel.sample(theta.size()).cuda()
            )
            theta.mul_(mul_)
            

def mlesigma(f_name, st_indx, end_indx):
    '''
    To estimate the equivalent standard deviation of log normal distribution from the I-V test data.
    In:
    - f_name: file name of the I-V test data file (*.csv)
    - st_indx: the start index of the effective data
    - end_indx: the end index of the effective data
    Out:
    '''
    
    ...
    
    return sigma_hat


def LCIS(seq):
    '''
    To calculate the Longest Conductance Increasing Subsequence (LCIS) in any input 1D sequence. 
    In:
    - seq: input sequence
    Out:
    - longest_start: start index of the LCIS
    - longest_end: end index of the LCIS
    '''
    
    ...
    
    return longest_start, longest_end


def calculate_smoothed_cmean(f_name):
    '''
    To return a smoothed time for any input 1D series data.
    In:
    - f_name: file name for an input sequence
    Out: 
    - smoothed: smoothed sequence ouput
    '''
    
    ...
    
    return smoothed






\end{lstlisting}


\section*{Supplementary Note 5: Details on Noise Injection on Different Networks}

\subsection*{Noise Injection using BayesMulti on PointPillar}
PointPillar is an efficient 3D object detection algorithm for point clouds \citep{lang2019pointpillars}. Compared with the previous classic PointNet architecture, it treats the point cloud as a cluster of columns whose positions are determined by x, y coordinates, and the eigenvector of each column contains important information about the point in the column, such as the maximum, the minimum, and average value of the height, as well as the number of points. In this way, the model can extract useful features from the raw point cloud data for subsequent object detection. The advantage of this architecture is its ability to efficiently process large amounts of 3D point cloud data while leveraging existing 2D convolutional neural network techniques for feature extraction and object detection. This allows PointPillar to outperform many other 3D object detection models in terms of performance and has significant advantages in computational efficiency. 
Specifically, the architecture of the PointPillar model is divided into the following steps: 

\begin{enumerate}
    \item Pillar Feature Net: This is the first layer of PointPillar and is responsible for converting 3D point cloud data into a 2D representation of the column feature. It first divides the 3D space into a set of fixed-size columns, then calculates the features of the point cloud data in each column (such as maximum, minimum, average, etc.), and generates a feature vector. This eigenvector can effectively encode the point cloud data in the column.
    \item 2D Convolution Layers: After converting 3D point cloud data into 2D column features, PointPillar uses a series of 2D convolutional layers to process these features and extract the high-level features from them. Since the column features are already organized into 2D, standard 2D convolution operations can be applied directly. This greatly simplifies the complexity of the model and allows the model to take advantage of existing 2D convolutional neural network techniques.
    \item Dense Head for 3D Object Detection: Finally, PointPillar uses a regression network to predict the bounding box of a 3D object. This regression network can predict the objects' position, size, and orientation in each bar. In this way, the model can generate a 3D bounding box that can be used to locate and identify objects in the environment.
\end{enumerate}

In our experiments, as shown in Figure \ref{fig: Archi PointPillar} we introduce noise-injecting layers into the Pillar Feature Net and 2D Convolution Layers.

\subsection*{Noise Injection using BayesMulti on Mason's CNN}
Mason's CNN architecture is designed to enhance the prediction of neutralizability for previously unseen antibodies \citep{mason2021optimization}. It achieves this by leveraging the extraction of local features from amino acid sequences.
The architecture consists of two CNN modules, each extracting relevant features specifically related to Ag and Ab. The extracted Ag and Ab features are subsequently fused. In addition, the architecture includes two independent linear layers that play a crucial role in the final prediction process. When applying our approach to Mason's CNN, we introduce noise-injection layers in the CNN modules as well as the last two linear layers. Figure \ref{fig: MasonCNN} illustrates the noise injection details on Mason's CNN architecture.

\subsection*{Noise Injection using BayesMulti on SweetNet}
SweetNet is a graph convolutional neural network (GCN) that can be employed for handling tasks related to polysaccharides \citep{burkholz2021using}. Its basic architecture consists of three graph convolutional layers and three linear layers, as illustrated in figure \ref{fig: SweetNet}. The data passes through the three graph convolutional layers successively, yielding three intermediate results. Each of these intermediate results is then subjected to graph max pooling and graph average pooling operations, resulting in three tensors respectively. After summing the three tensors mentioned above, the final output is obtained through three linear layers in sequence, which can be used for tasks such as glycan classification. When applying our approach on SweetNet, we introduce noise before three graph convolutional layers and the first two linear layers.


\newpage
\renewcommand\thefigure{S\arabic{figure}}
\setcounter{figure}{0}
\begin{figure}[H]
    \centering
    \includegraphics[width=0.9\linewidth]{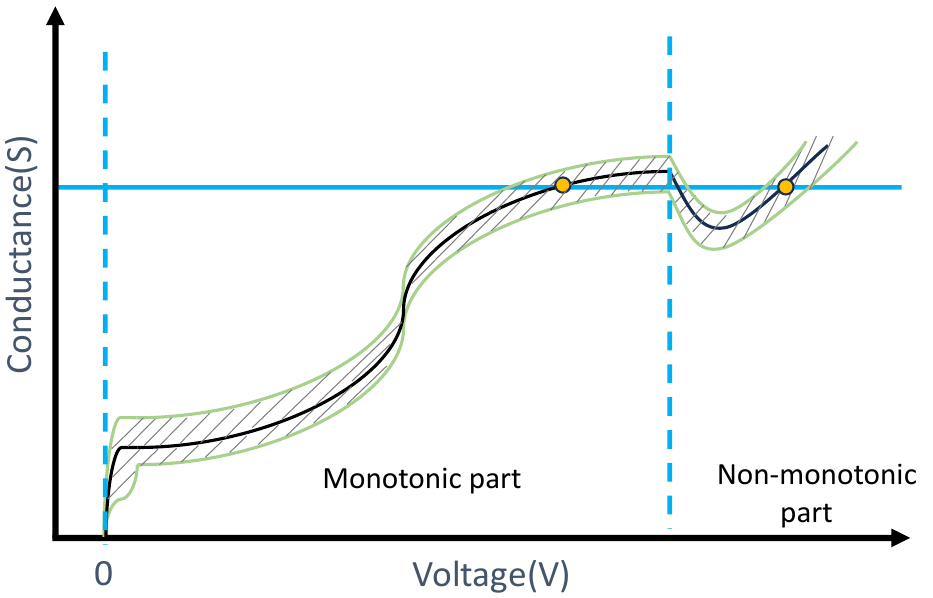}
    \caption{Illustration of hardware non-idealities. The black line and the green lines denote the mean conductance and deviations respectively, across different cycles of measurements. The dashed blue lines segment the monotonic and the non-monotonic parts of the I-V curve. The yellow points in the monotonic and non-monotonic parts correspond to the same conductance level at different sweeping voltages.}
    \label{fig:hwnoise_illustrate}
\end{figure}

\begin{figure}[H]
    \centering
    \includegraphics[width=\linewidth]{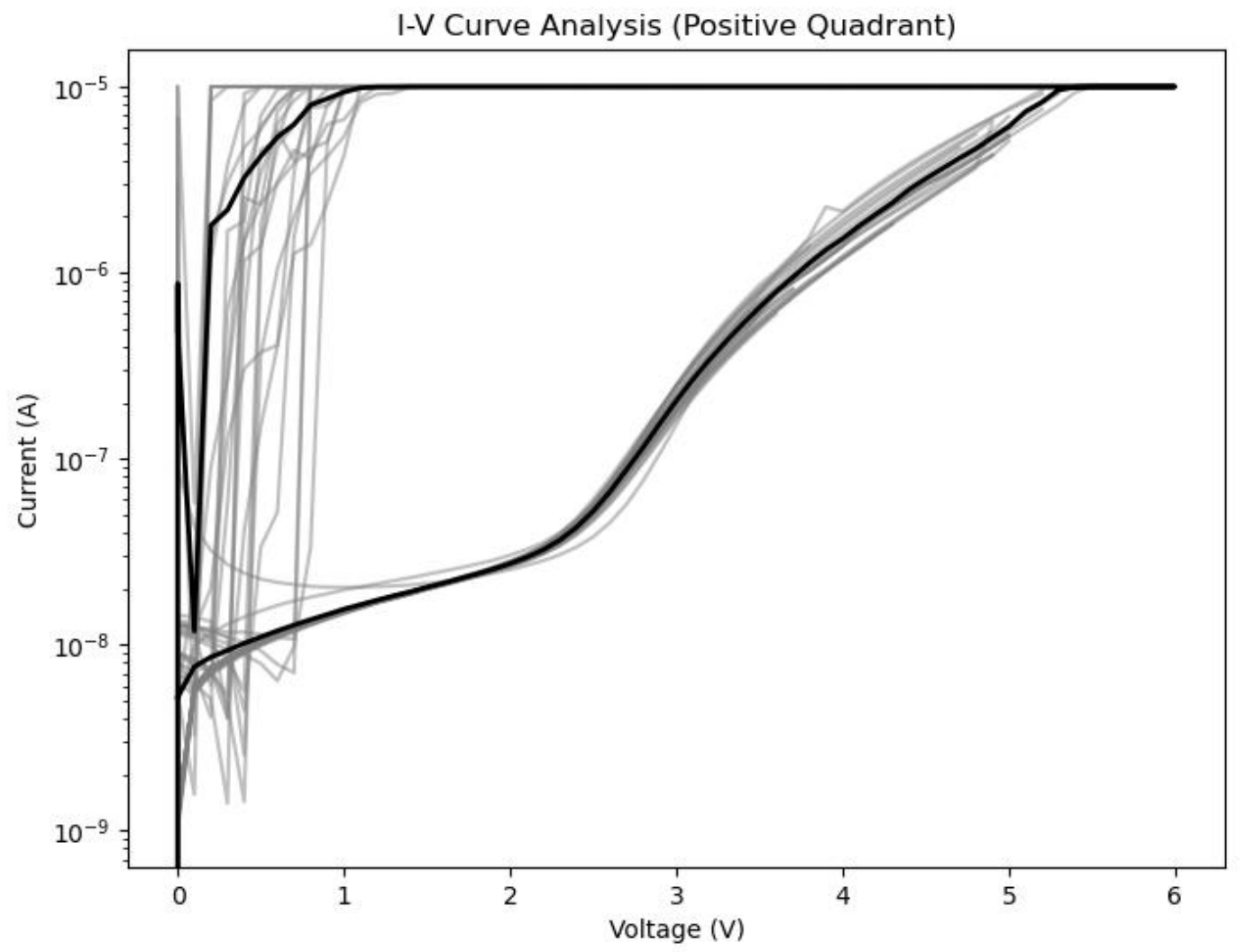}
    \caption{The positive quadrant of I-V characteristics of a well-performed perovskite NW-based memristor. The grey curves are from raw data and the bold black curve represents the average performance.}
    \label{fig:good_device}
\end{figure}

\begin{figure}[H]
    \centering
    \includegraphics[width=\linewidth]{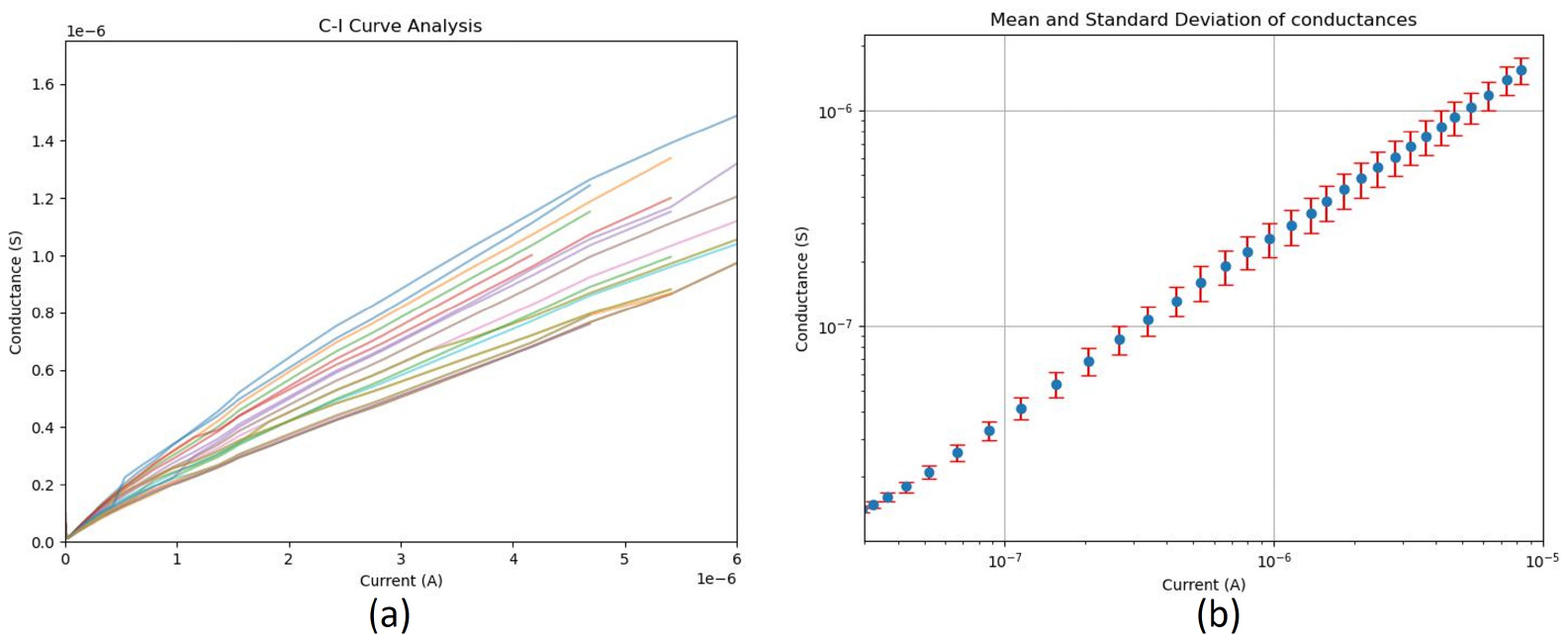}
    \caption{(a) The conductance vs. current of different cycles of a well-performed memristor. (b) the mean and standard deviation of conductance of the well-performed memristor at different current levels. The scatters represent the mean values and the error bars represent the standard deviation.}
    \label{fig:mean_deviation_good}
\end{figure}

\begin{figure}[H]
    \centering
    \includegraphics[width=\linewidth]{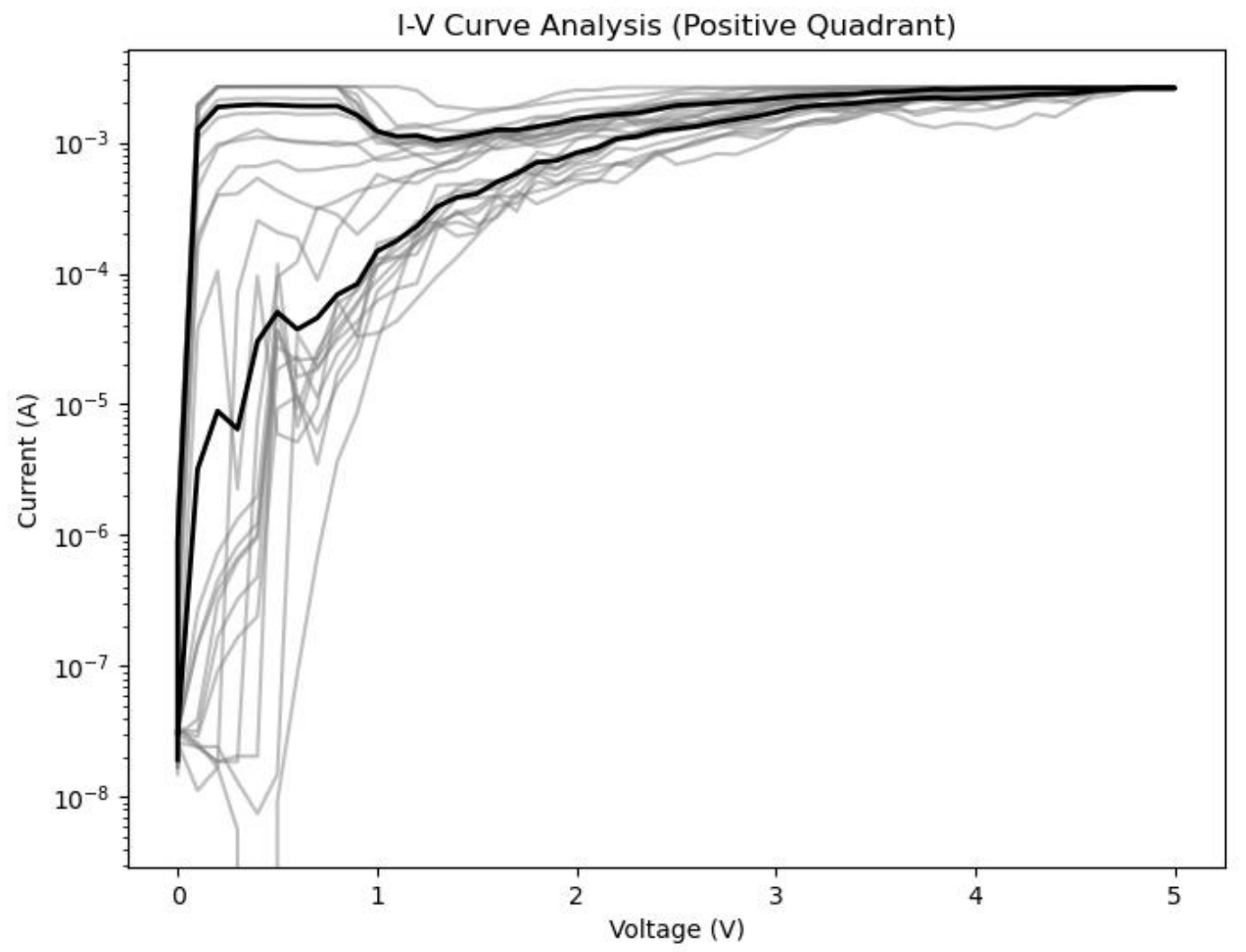}
    \caption{The positive quadrant of an I-V characteristics of a poorly performed perovskite NW-based memristor. The grey curves are from raw data and the bold black curve represents the average performance.}
    \label{fig:0.7um_device}
\end{figure}

\begin{figure}[H]
    \centering
    \includegraphics[width=\linewidth]{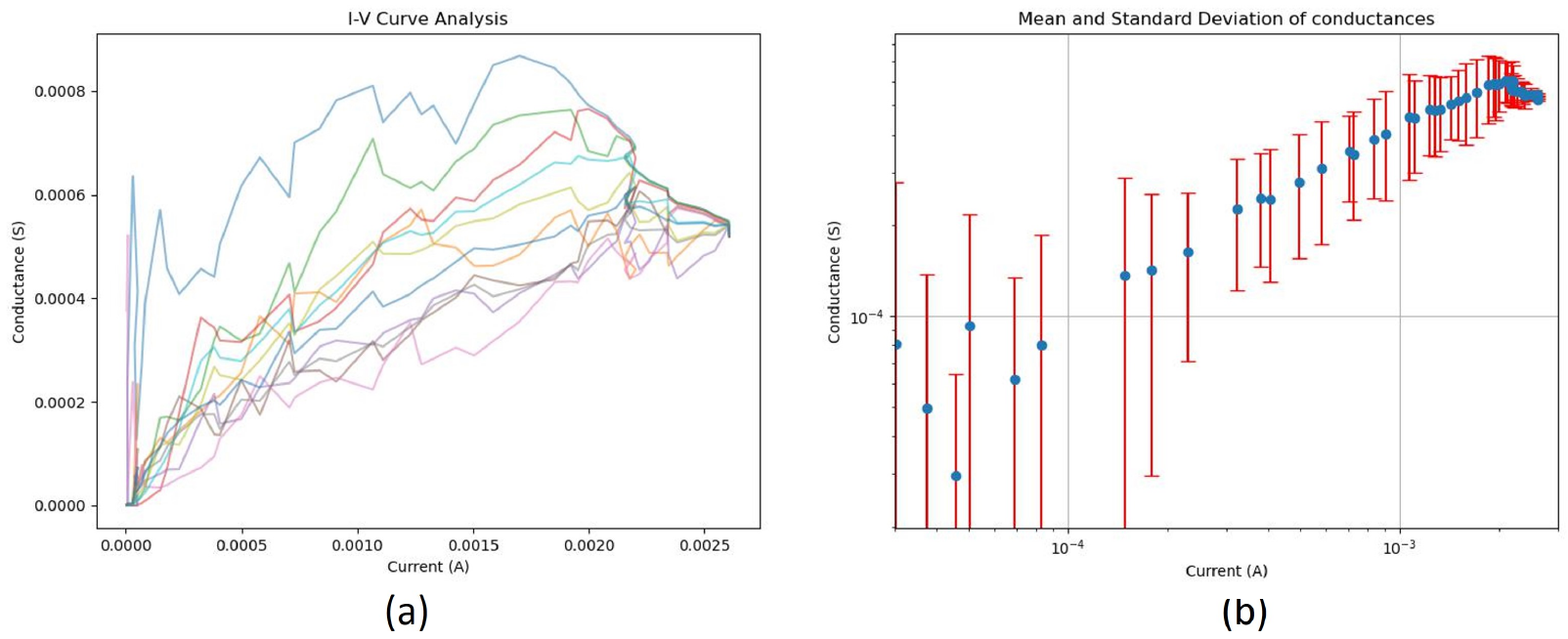}
    \caption{(a) The conductance vs. current of different cycles of a poorly-performed memristor. (b) the mean and standard deviation of conductance of the poorly-performed memristor at different current levels. The scatters represent the mean values and the error bars represent the standard deviation.}
    \label{fig:0.7um mean_deviation_good}
\end{figure}

\begin{figure}[H]
    \centering
    \includegraphics[width=0.8\linewidth]{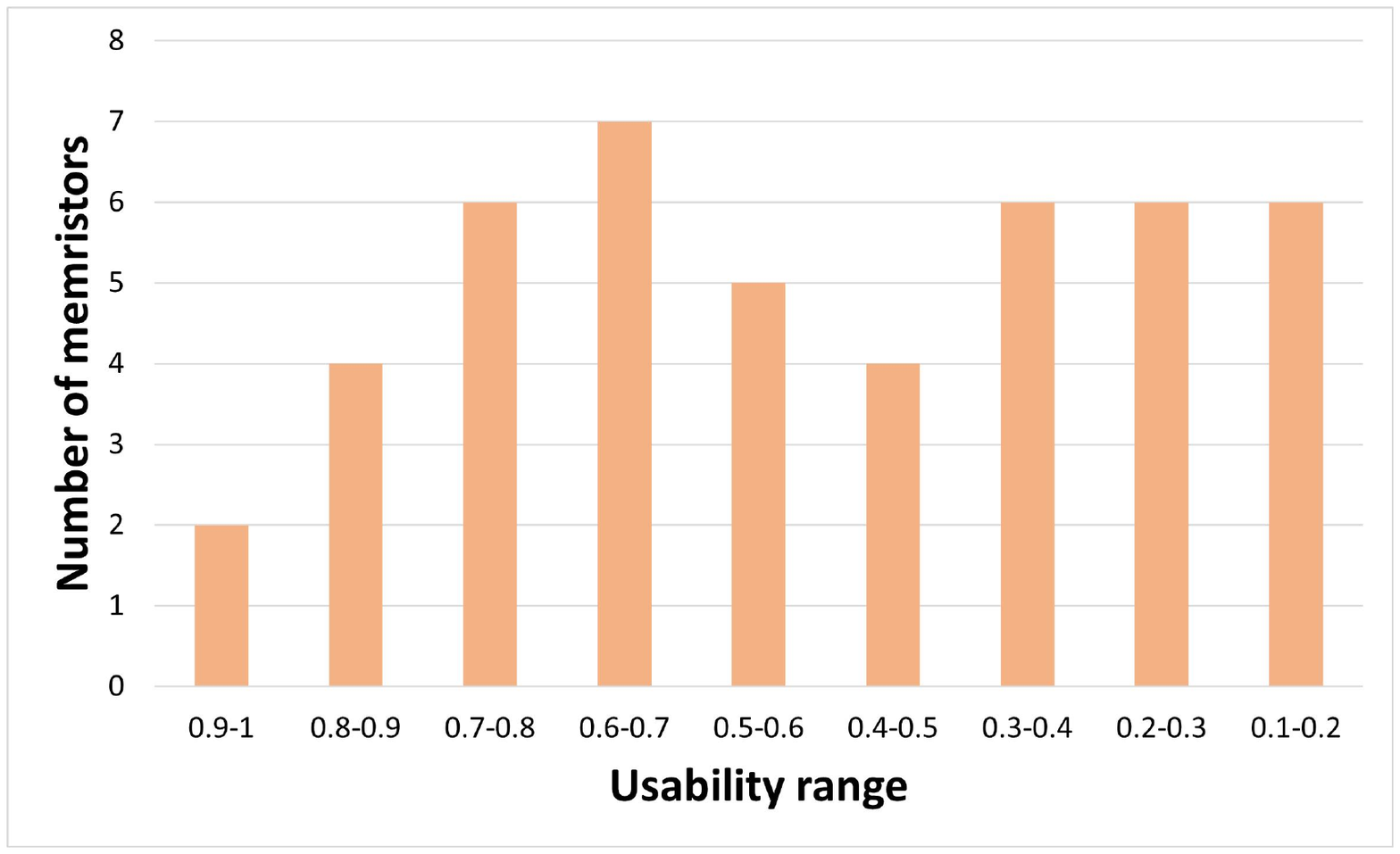}
    \caption{The number of fabricated perovskite NW-based memristors in a typical usability range.}
    \label{fig:usability_distribution}
\end{figure}

\begin{figure}[H]
    \centering
    \includegraphics[width=\linewidth]{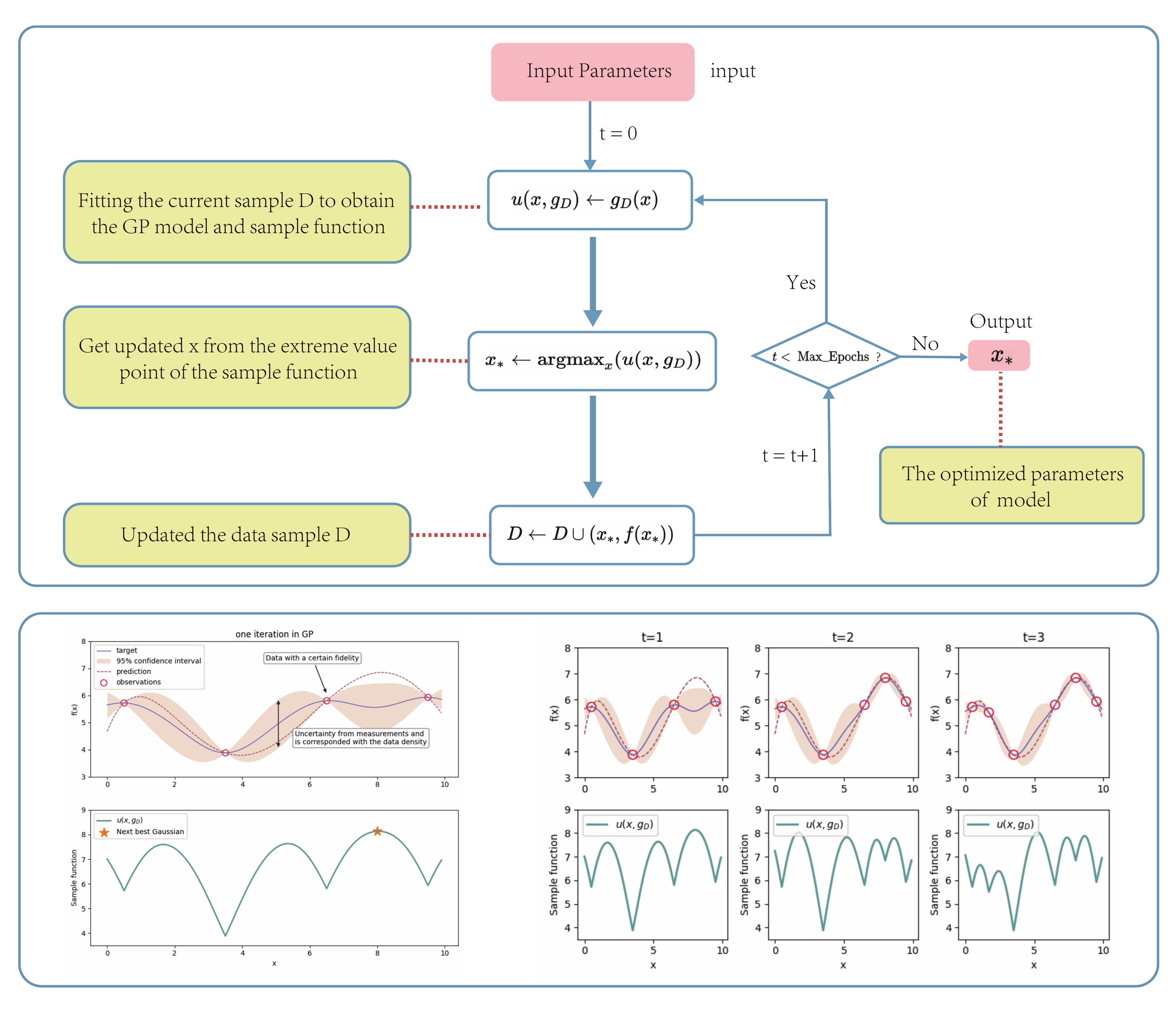}
    \caption{The process of Bayesian Optimization}
    \label{fig:BO_process}
\end{figure}

\begin{figure}[H]
    \centering
    \includegraphics[width=\linewidth]{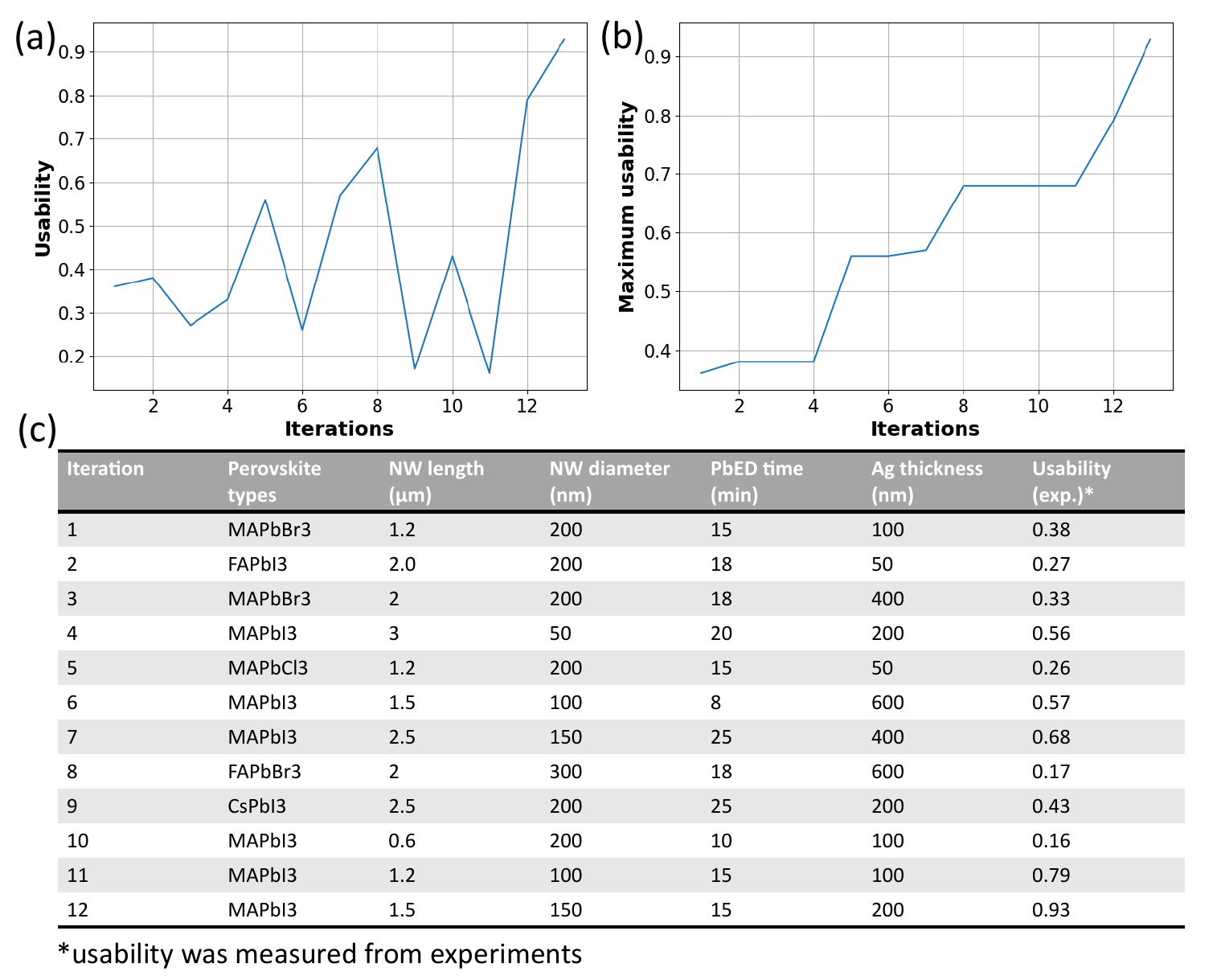}
    \caption{(a) usability vs. iterations (b) Maximum usability vs. iterations in Bayesian fabrication optimization process. (c) The fabrication conditions chosen by the BO process and the yielded usability values in each iteration.}
    \label{fig:BO fabricate}
\end{figure}


\begin{figure}[H]
    \centering
    \includegraphics[width=0.8\linewidth]{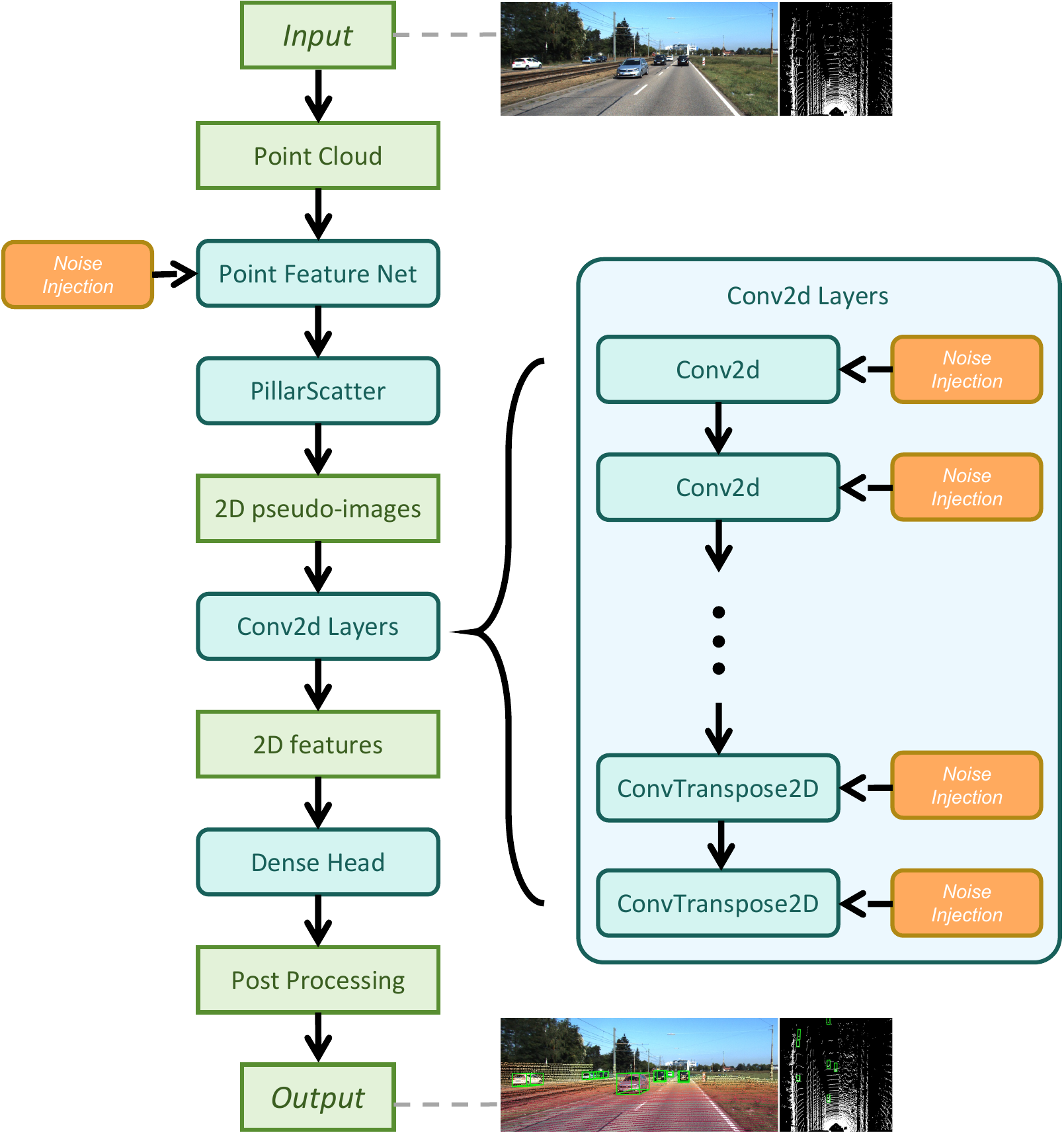}
    \caption{The architecture PointPillar for the autonomous driving task. Noise-injecting layers are applied on the Pillar Feature Net and 2D Convolution Layers.}
    \label{fig: Archi PointPillar}
\end{figure}

\begin{figure}[H]
    \centering
    \includegraphics[width=0.6\linewidth]{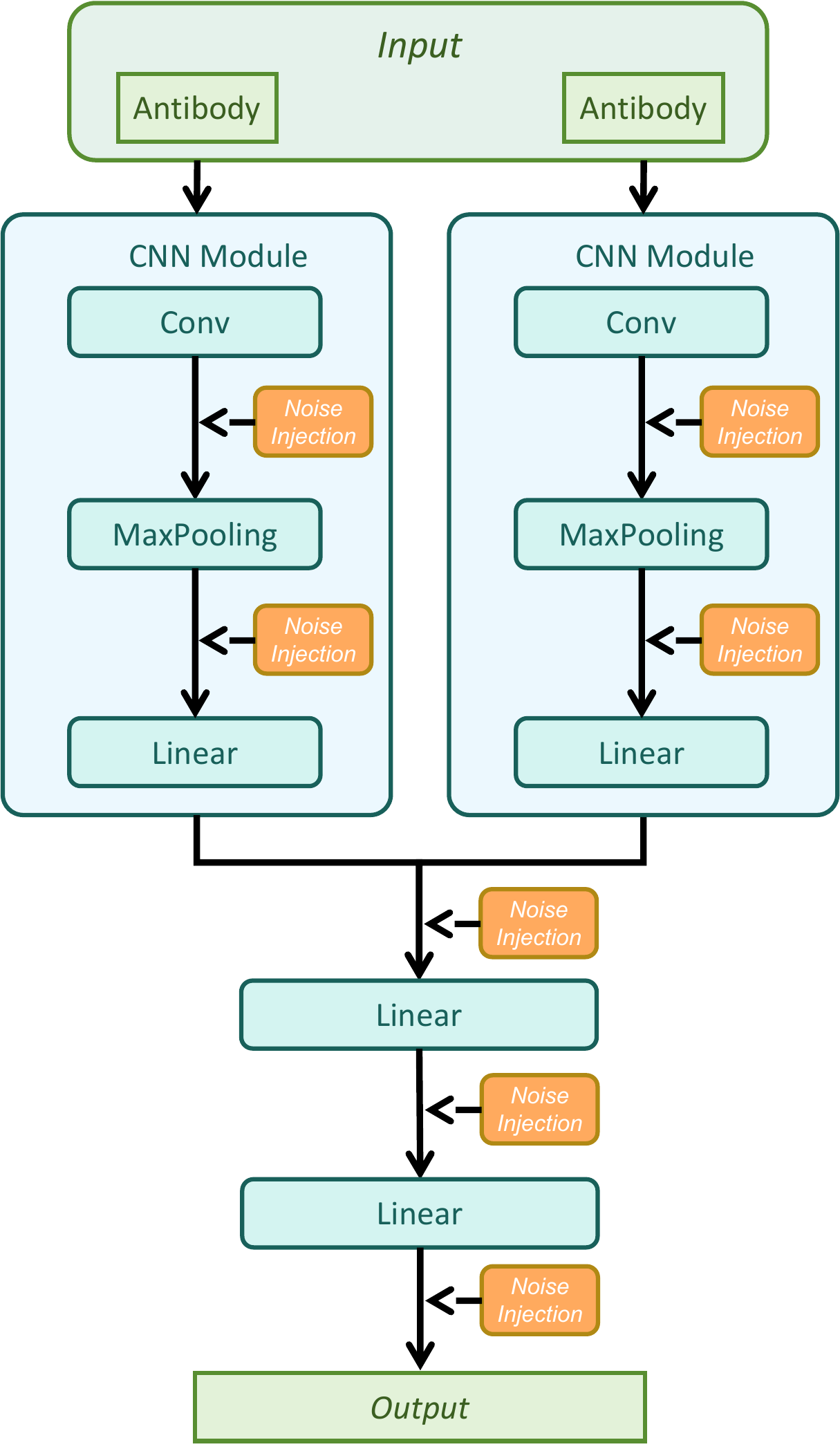}
    \caption{The architecture of Mason's CNN. Noise-injecting layers are applied to CNN modules and the last two linear layers.}
    \label{fig: MasonCNN}
\end{figure}

\begin{figure}[H] 
    \centering
    \includegraphics[width=0.65\linewidth]{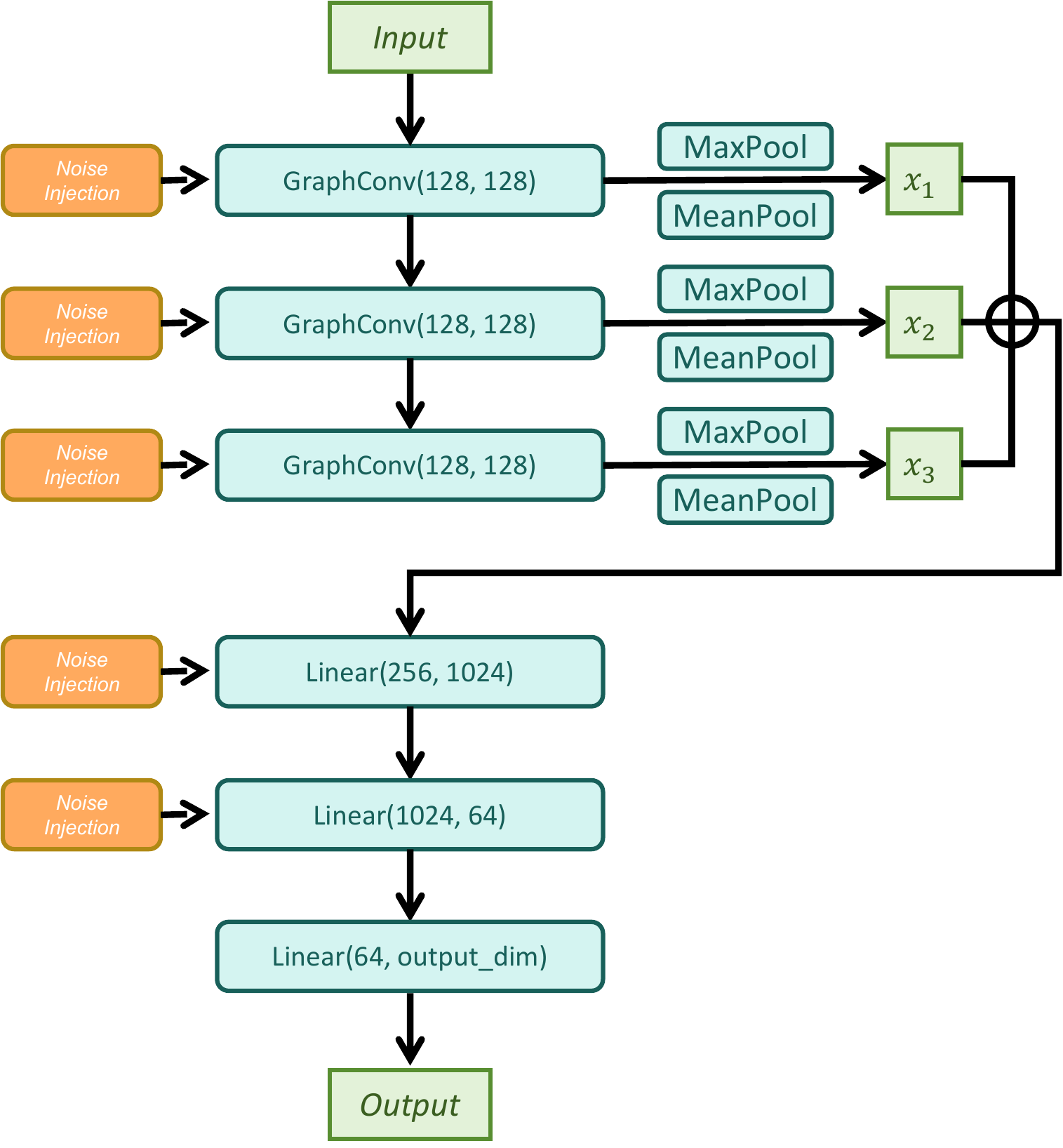}
    \caption{The architecture of SweetNet. Noise-injecting layers are applied to graph convolutional layers and the first two linear layers.}
    \label{fig: SweetNet}
\end{figure}

\newpage
\def\bibsection{\section*{Supplementary Reference}}
\bibliography{references}

\end{appendices}


\end{document}